\pdfoutput=1
\documentclass{article}

\usepackage[main, final]{neurips_2026}

\usepackage[utf8]{inputenc}
\usepackage[T1]{fontenc}
\usepackage{hyperref}
\usepackage{url}
\usepackage{booktabs}
\usepackage{amsfonts}
\usepackage{nicefrac}
\usepackage{microtype}
\usepackage{xcolor}

\usepackage{graphicx}
\usepackage{subcaption}
\usepackage[ruled,vlined]{algorithm2e}
\usepackage{amsmath} 
\usepackage{multirow}
\usepackage{wrapfig}

\title{CECAR : Cache \& Expert Co-Aware Routing Accelerates On-Device Inference of MoE LLMs}

\author{
 \textbf{Byeongju Kim}$^{1}$\thanks{Equal contribution.}\quad
 \textbf{Hoonki Lee}$^{2}$\footnotemark[1]\quad
 \textbf{Byungjun Kim}$^{3}$\quad
 \textbf{Inyul Ra}$^{3}$\quad
 \textbf{Yongjin Kim}$^{3}$ \\
 \textbf{Yongseong Lee}$^{3}$\quad
 \textbf{Sangyeob Kim}$^{3}$\thanks{Corresponding author.} \\[0.9em]
 $^{1}$KAIST \quad
 $^{2}$Agency for Defense Development \quad
 $^{3}$Yonsei University \\[0.4em]
 \texttt{byeongju.kim@kaist.ac.kr} \quad
 \texttt{hoonki2002@gmail.com} \quad
 \texttt{malmong@konkuk.ac.kr} \\
 \texttt{\{inyul0703, kyw189, lys7523, sangyeob.kim\}@yonsei.ac.kr}
}

\begin{document}

\maketitle

\begin{abstract}
Despite growing interest in on-device LLM deployment, large-scale Mixture-of-Experts (MoE) models remain impractical on resource-constrained devices, as sparse expert activation still requires all expert weights to be memory-resident. To address this, we identify three key observations for on-device MoE inference: (a) cache hit rates are fundamentally bounded even under Belady’s optimal policy; (b) these bounds are insufficient for low-latency on-device inference; and (c) prior cache-aware routing based on binary cache presence can induce routing instability. Based on these insights, we propose CECAR (Cache \& Expert Co-Aware Routing), a MoE inference system that jointly considers expert routing and cache management. Unlike prior cache-aware routing methods, CECAR uses an ML-based cache policy to predict the expected reuse distance of each expert and prioritizes routing and execution toward experts with smaller predicted reuse distances. On the Qwen3-30B-A3B model, CECAR achieves 12.5 tokens/s on a consumer-grade GPU with only 16 GB of VRAM, where non-resident experts are fetched from SSD, delivering a 4.5× speedup over conventional MoE inference and exceeding the decode speed of baselines that keep the entire model fully resident in DRAM. Code is available at \url{https://github.com/hoonki2002/CECAR}.

\end{abstract}

\section{Introduction}

Large language models (LLMs) have achieved significant advances, but their deployment on user-centric devices remains challenging due to strict memory and latency constraints. While on-device inference is increasingly preferred for privacy, latency, and cost reasons \citep{li2024personalllmagentsinsights}, limited hardware resources restrict executable model size to a few billion parameters (e.g., Mistral 7B \citep{jiang2023mistral7b}) \citep{xu2024ondevicelanguagemodelscomprehensive}. This trend is already reflected in commercial systems such as Microsoft Copilot and Samsung Galaxy devices, which increasingly provide on-device AI functionalities for tasks including summarization, translation, and voice interaction.

\begin{wrapfigure}{r}{0.40\linewidth}
  \centering
  \includegraphics[width=\linewidth]{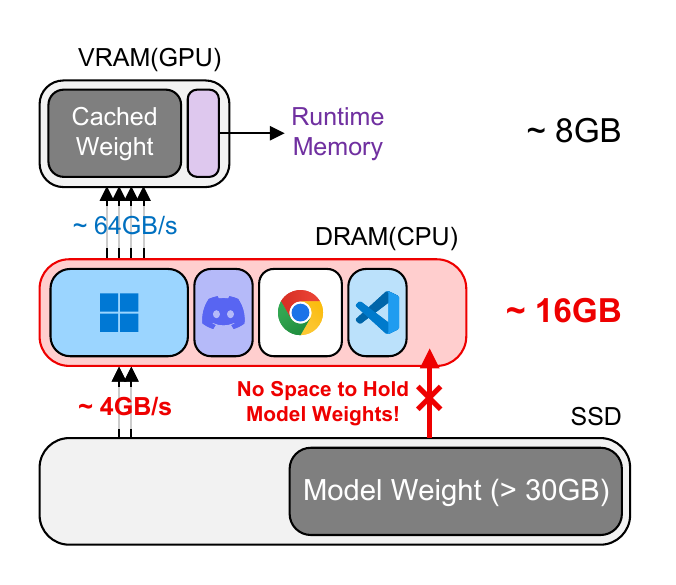}
  \caption{Memory bottlenecks for MoE inference on consumer devices.}
  \label{fig:overview}
\end{wrapfigure}

Mixture-of-Experts (MoE) architectures \citep{shazeer2017} mitigate computation by activating only a subset of experts per token, enabling large model capacity with reduced per-token cost. However, the total parameter footprint of all experts remains large, introducing a critical bottleneck in on-device environments. As illustrated in Figure~\ref{fig:overview}, the limited DRAM capacity of typical devices (8–16 GB) prevents most experts from residing in memory, forcing them to be stored in lower memory tiers such as SSDs \citep{alizadeh-etal-2024-llm}. Consequently, dynamically loading top-$k$ experts incurs frequent memory transfers, making memory access latency the dominant factor limiting throughput (e.g., $\sim$10 tokens/s). In particular, even a small number of uncached experts can dominate decoding time, highlighting the importance of efficient cache utilization. As a result, MoE architectures alone are insufficient to enable practical on-device inference.

Prior work addresses this challenge through expert pruning and cache-aware routing. Pruning methods such as DES \citep{lu-etal-2024-experts} and ODP \citep{huang2025mixture} reduce activated experts, improving efficiency at the cost of reduced model capacity and degraded accuracy on benchmarks such as MATH \citep{hendrycks2021measuringmathematicalproblemsolving} and HumanEval \citep{chen2021evaluatinglargelanguagemodels}. Cache-aware routing methods, such as MoCCE \citep{skliar2025mixture}, target on-device settings by improving cache hit rates through preferential selection of cached experts.

However, these approaches fail to address key challenges of on-device inference. Even with pruning, many experts must still be repeatedly loaded from SSDs, and existing methods do not effectively exploit temporal locality by reusing cached experts. Moreover, pruning requires aggressive reduction under small cache budgets, further degrading accuracy, while cache-aware routing relies on binary cache signals and cannot distinguish eviction priorities among cached experts.
As a result, existing methods remain insufficient to simultaneously achieve high efficiency and stable accuracy under constrained cache capacity.

To address these challenges, we propose CECAR (Cache \& Expert Co-Aware Routing), which jointly considers expert cache residency and predicted eviction likelihood under constrained cache capacity. CECAR integrates cache residency and predicted eviction likelihood into the routing process, maximizing inference speed while minimizing accuracy degradation in on-device environments. It is composed of three tightly coupled components:

\textbf{Cache-Aware Routing.}
Cache residency is explicitly incorporated into the routing process, 
biasing expert selection toward those already present in VRAM. 
Unlike conventional MoE inference that strictly follows router logits, 
this mechanism reduces expensive expert loading from lower memory tiers 
(e.g., SSD), which dominates token-level latency in constrained environments. 
Furthermore, it applies cache-conditioned pruning during expert selection, 
skipping uncached low-priority experts. 
Importantly, cache awareness is applied during expert selection without 
altering the final routing weights, preserving the model’s original behavior.

\textbf{Expert-Aware Routing.} 
Semantic preference of the original router is preserved by maintaining 
alignment with its ranking over experts. Rather than replacing routing decisions based solely on cache status, it ensures that selected experts remain consistent with the model’s learned routing distribution. To achieve this, CECAR applies margin-based adjustments as controlled logit perturbations within the top-ranked candidates, allowing flexible reordering without deviating significantly from the original routing. This design maintains semantic consistency while preventing routing instability, avoiding the cascading errors observed in binary cache-aware approaches during autoregressive decoding.

\textbf{ML-Based Cache Policy.}
A lightweight learning-based cache policy is used to predict the reuse 
likelihood of each expert. The policy leverages normalized recency and 
frequency signals to produce fine-grained eviction scores, enabling differentiation 
between frequently reused (“hot”) and rarely reused (“cold”) experts. 
These scores are directly integrated into the routing stage as eviction-aware 
bonus terms, allowing CECAR to jointly optimize cache replacement and expert 
selection under varying cache sizes and workloads.

\begin{figure*}[t]
  \centering
  \includegraphics[width=0.90\linewidth]{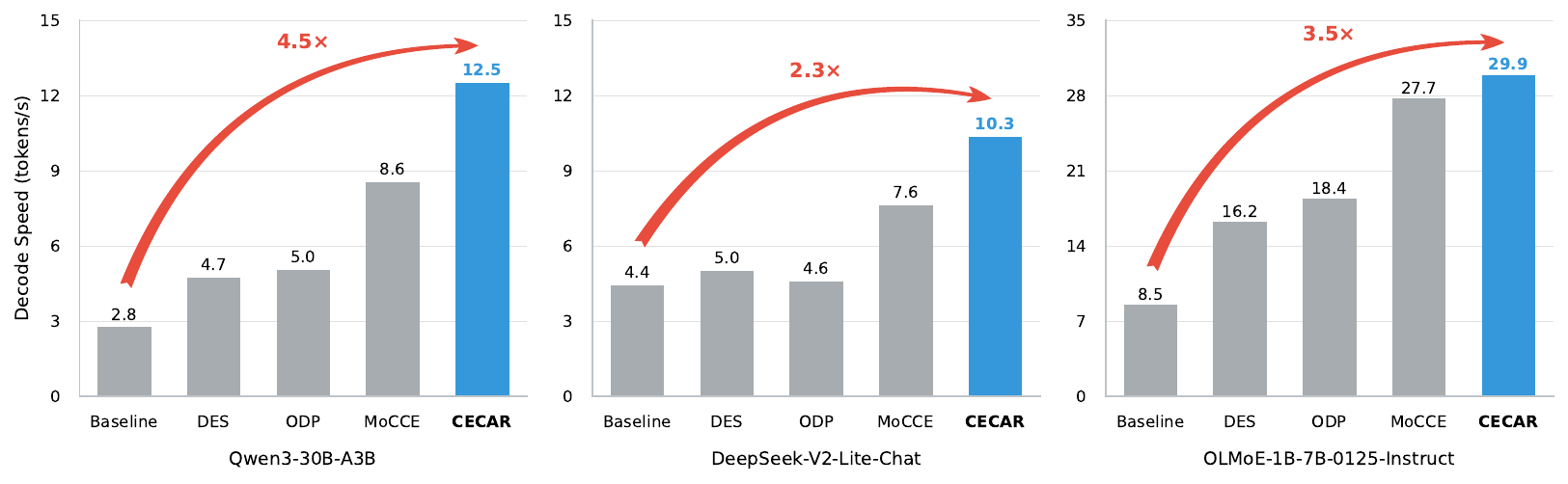}
  \caption{Decoding Speed (Tokens/s) Across MoE Models under Different Routing Strategies. }
  \label{fig:cecarperf}
\end{figure*}

We demonstrate that, in an on-device environment with 16~GB VRAM and 1~GB system DRAM, our approach enables inference of a 30B-parameter LLM at 12.5 tokens/s using SSD-based storage (Figure~\ref{fig:cecarperf}), achieving a 4.5$\times$ speedup over conventional MoE inference and outperforming DES (1.7$\times$) and MoCCE (3.1$\times$). 
Moreover, compared to a baseline that executes the same MoE model on the same GPU (16~GB VRAM) with all expert parameters (approximately 56.7~GB) fully resident in system DRAM, achieving 7 tokens/s, our approach delivers a 1.8$\times$ higher inference throughput while reducing the total memory footprint (VRAM + system DRAM) by approximately 78.8\%.
These results demonstrate that 30B-scale LLMs, which typically require more than 60~GB of system DRAM, can be executed at practical speeds even in memory-constrained on-device environments, and further suggest the potential for scaling to even larger models.

\section{Related Works}

\subsection{Expert Caching}

To overcome memory capacity limitations in MoE inference systems, prior work has proposed caching recently used experts. Mixtral-Offloading\citep{eliseev2023fastinferencemixtureofexpertslanguage} was the first to apply LRU caching to expert layers in MoE models, achieving a 2.2× speedup compared to naïve offloading approaches designed for dense models. FlashMoE\citep{kim2026flashmoereducingssdio} further improved accuracy by leveraging a calibration set based on Belady’s optimal policy\citep{5388441}, achieving up to a 2.6× speedup over heuristic cache policies such as LRU and LFU. These studies demonstrate that expert caching is feasible in MoE inference systems and can yield substantial speed improvements when properly managed. However, they do not fully exploit a key property of MoE models—that routing to alternative experts, even when not exact, does not necessarily incur significant accuracy degradation.

\subsection{Expert Pruning}
Expert pruning techniques have also been proposed, in which inference does not strictly follow the experts learned during training, but instead selectively omits or replaces certain experts. In MC-MoE\citep{huang2025mixture}, Dynamic Expert Skipping (DES) dynamically skips experts with low contribution among the selected experts, while Online Dynamic Pruning (ODP) further mitigates accuracy degradation by incorporating attention-based key token protection. These approaches require prior calibration or threshold tuning to minimize accuracy loss after pruning, which can introduce additional overhead in terms of latency and memory usage in practical on-device environments. More importantly, these studies do not consider cache locality during the routing stage, and therefore cannot be regarded as optimized for on-device settings.

\subsection{Efficient MoE Inference}
Beyond caching and pruning, a variety of techniques have been proposed to accelerate MoE inference by optimizing hardware resource utilization or improving model representations. Fiddler\citep{kamahori2024fiddler} dynamically determines where computation should be performed for each token and expert, alleviating CPU–GPU communication bottlenecks and achieving up to a 1.26× inference speedup. HOBBIT\citep{tang2024hobbitmixedprecisionexpert} reduces both memory footprint and computational cost of expert layers by applying mixed-precision techniques based on expert importance. Beyond these, recent expert-offloading systems intervene at three distinct points of the pipeline.
(i) \emph{Exact-routing caching and prefetching}: MoE-Infinity\citep{xue2025moeinfinityefficientmoeinference} targets single-user, batch-size-one inference on personal machines with a sparsity-aware expert cache and activation-trace-based prefetching, and FineMoE\citep{yu2025taminglatencymemorytradeoffmoebased} guides asynchronous prefetching and eviction per request using semantic and trajectory signals. These systems leave the router's decision untouched and hide I/O by predicting which experts will be needed.
(ii) \emph{Scheduling and deferral}: KTransformers\citep{10.1145/3731569.3764843} keeps selected experts in host memory, runs them with optimized CPU kernels, and defers the outputs of low-score experts to a later layer to overlap CPU and GPU work; it assumes enough host DRAM for the experts rather than trying to reduce it.
(iii) \emph{Cache-aware expert selection}: CECAR adjusts the routing decision itself so that fewer non-resident experts are requested in the first place.
A system that keeps exact routing cannot exceed the hit rate of Belady's policy on the original routing; CECAR is not bound by it, and the two approaches can be combined (Appendix~\ref{appendix:offloading-comparison}).

\begin{figure*}[t]
  \centering
  \begin{subfigure}[t]{0.48\linewidth}
    \centering
    \includegraphics[width=\linewidth]{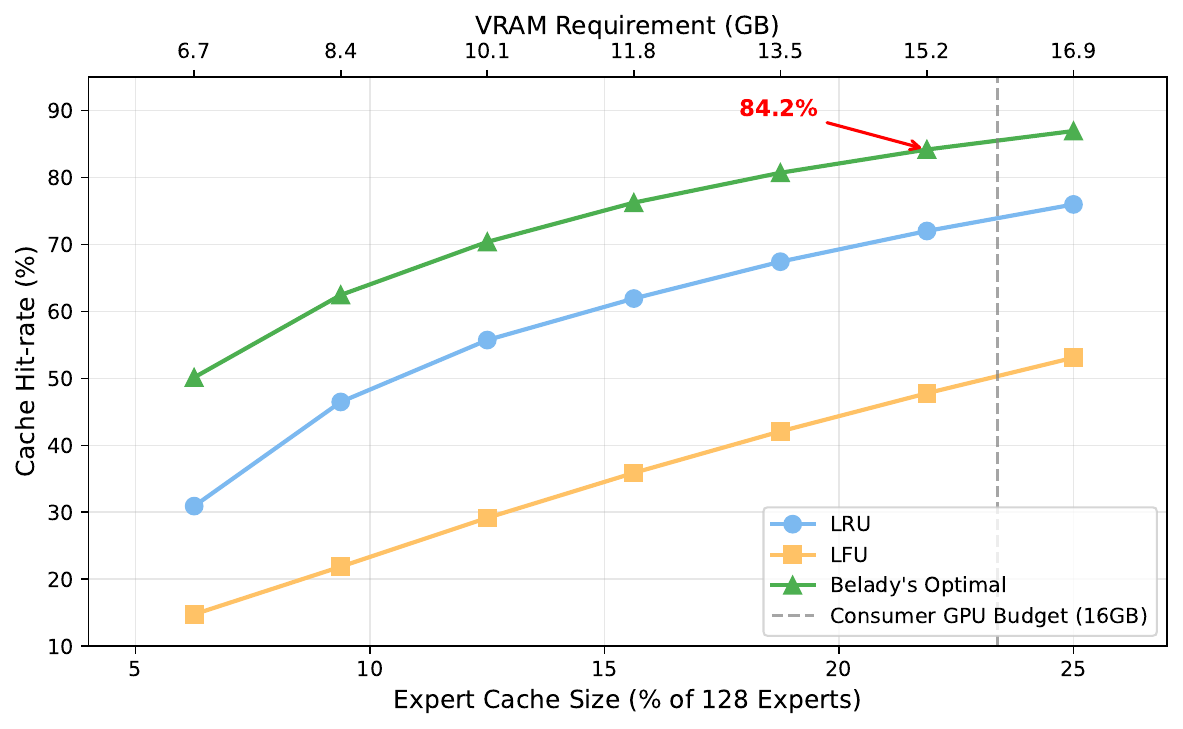}
    \caption{Achieved hit rates for different expert cache sizes.}
    \label{fig:belady}
  \end{subfigure}
  \hfill
  \begin{subfigure}[t]{0.48\linewidth}
    \centering
    \includegraphics[width=\linewidth]{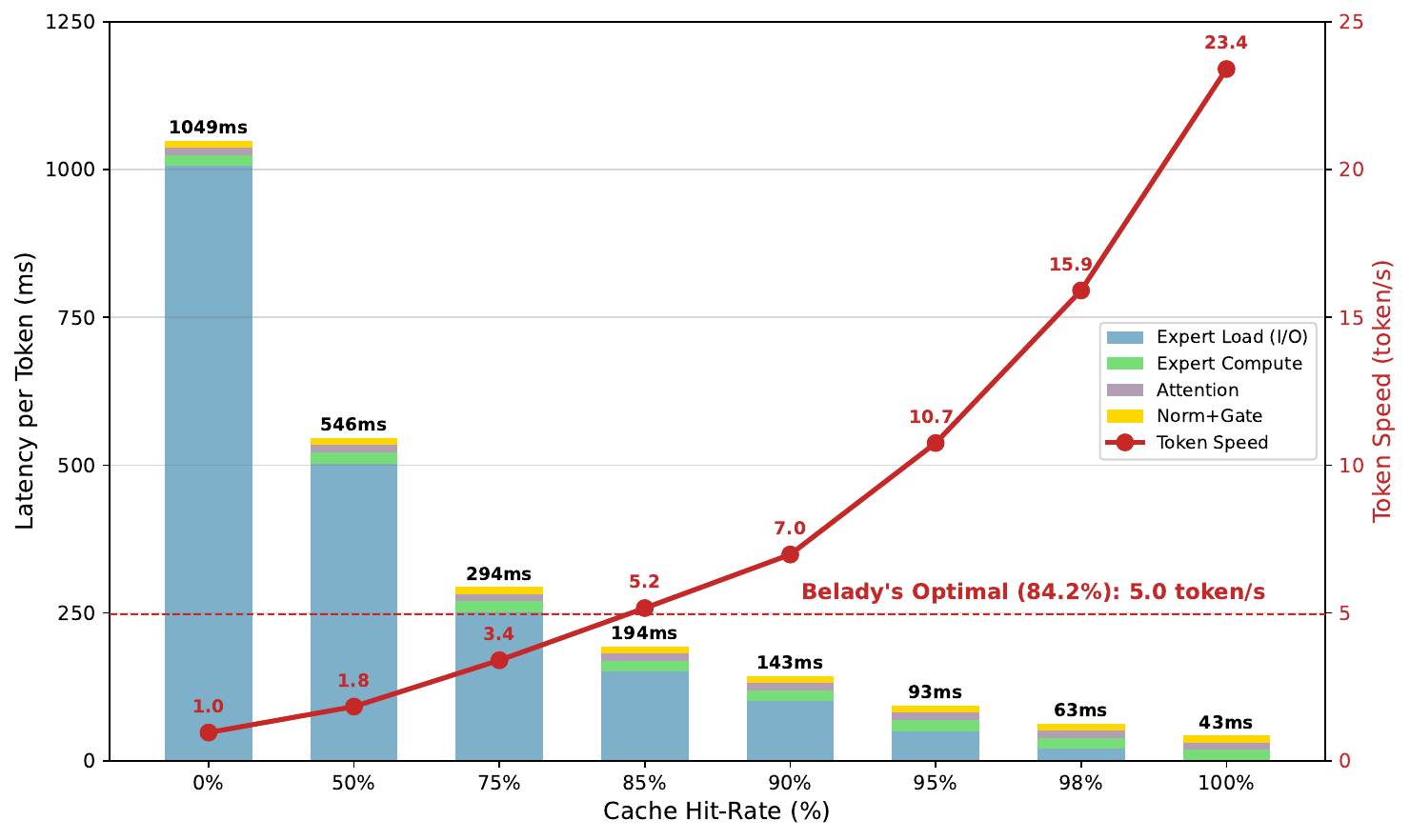}
    \caption{Impact of cache hit rate on latency and token throughput.}
    \label{fig:cache-hitrate}
  \end{subfigure}
  \caption{Latency and memory implications of cache hit rate and expert cache size.}
  \label{fig:cache-overview}
\end{figure*}

\section{Motivation}
When performing MoE LLM inference in on-device environments, limited memory capacity prevents all experts from being kept resident in memory. For example, under a setting where only 24 out of 128 experts can be stored in memory, conventional cache replacement policies such as Least Recently Used (LRU) and Least Frequently Used (LFU) exhibit substantially limited average layer-wise cache hit rates, as illustrated in Figure~\ref{fig:belady}.

Using the Qwen3-30B-A3B\citep{yang2025qwen3technicalreport} model, we measured cache hit rates across multiple datasets during inference and observed that the average hit rates of LRU and LFU are only 67\% and 45\%, respectively. This is approximately 10–30 percentage points lower than the hit rate of 80\% achieved by the ideal Belady’s optimal policy under the same conditions.
 Similar patterns were observed for the OLMoE-1B-7B-0125-Instruct\citep{muennighoff2025olmoe} and DeepSeek-V2-Lite-Chat\citep{deepseekai2024deepseekv2strongeconomicalefficient} models, and repeated experiments across diverse datasets consistently showed that LRU/LFU-based approaches yield significantly lower hit rates than Belady’s optimal policy.

A more critical issue is that even when assuming Belady’s optimal policy, cache hit rates remain insufficient in practical on-device settings with limited cache capacity. For the Qwen3-30B-A3B model, when only 24 out of 128 experts can be kept in memory, the average layer-wise hit rate reaches only 84.2\% even under the ideal Belady policy (Figure~\ref{fig:belady}). As shown in Figure~\ref{fig:cache-hitrate}, this hit rate still results in an I/O-dominated execution, where expert loading accounts for over 75\% of the per-token latency and limits the achievable decoding speed to around 5 tokens/s. Therefore, further improvements to cache replacement alone are fundamentally insufficient, motivating a routing-stage approach that explicitly accounts for cache locality.

However, cache-aware routing that relies solely on the binary status of cache presence (hit or miss) can introduce significant challenges in memory-constrained environments. To better understand the limitations of such approaches under tight memory constraints, we analyze the behavioral impact of routing-stage approximation. While prior work\citep{skliar2025mixture} reports minimal degradation in perplexity, we find that cache-aware routing based on binary cache presence can induce a stickiness effect, where certain experts persist in the cache despite limited reuse. Under MoCCE-style routing, 6.1\% of cache residency periods exhibit prolonged idle behavior (remaining in cache for over 20 tokens with fewer than five uses), with idle residency extending up to 53 tokens, indicating a long-tail risk of cache lock-in.

This cache-induced bias also affects routing stability during generation. In free-running decoding after cache warm-up, small routing perturbations introduced by binary cache-aware routing are amplified through autoregressive generation, causing the output to diverge from the exact-routing baseline after 23 tokens on average, with only 6.8\% token-level agreement. Although such divergence does not necessarily lead to immediate benchmark degradation, it reveals an instability where cache bias can cascade over time. These observations suggest that effective cache-aware routing must go beyond binary cache presence and selectively account for expert reuse characteristics to improve cache locality while preserving semantic stability.

Accordingly, the goal of this work is to leverage the memory constraints and expert flexibility observed above to substantially improve the inference efficiency of MoE-based LLMs in highly resource-constrained on-device environments. To this end, we propose CECAR, a novel routing policy that jointly accounts for cache state and expert reuse status, overcoming the structural limitations of existing exact-routing-based MoE inference methods.

\begin{figure*}[t]
  \centering
  \includegraphics[width=0.69\linewidth]{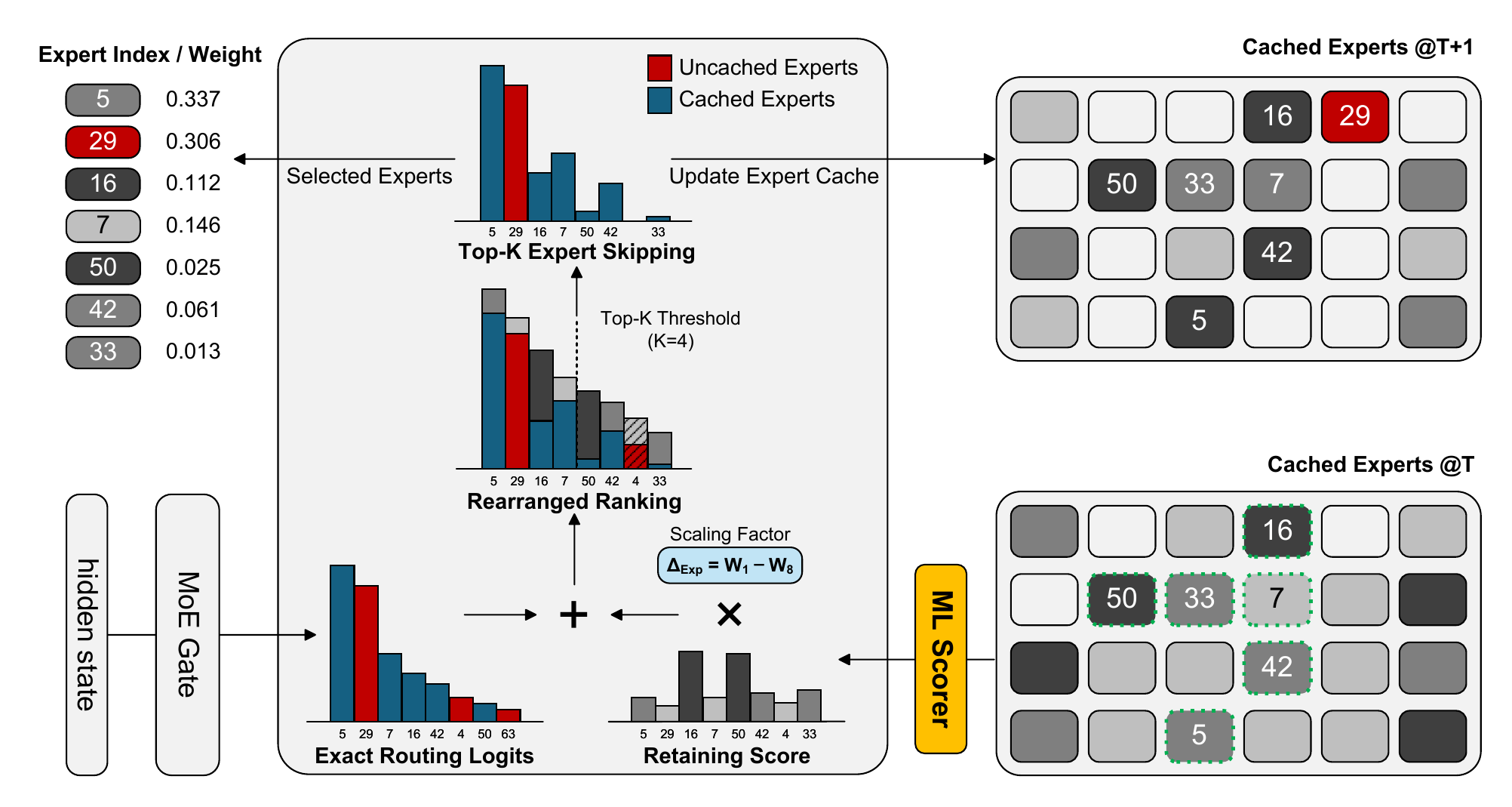}
  \captionsetup{justification=centering}
  \caption{Illustration of the CECAR Routing Algorithm. Darker colors denote experts that CECAR expects to be routed in the near future.}
  \label{fig:cecar-routing}
\end{figure*}

\section{Method}

\subsection{Preliminary}
The CECAR inference system targets a fine-grained Mixture-of-Experts (MoE) architecture. Each MoE layer consists of 64 to up to 128 experts, and 6 $\sim$ 8 experts are selected per token. This architecture poses a practical challenge in that, due to memory constraints, it is difficult to keep all experts resident in memory. Conventional heuristic-based exact caching policies (e.g., LRU and LFU) are simple and deterministic, but fail to capture the nonlinear expert selection patterns across tokens, resulting in limited cache hit rates. Accordingly, we propose a pre-trained, learning-based cache policy that can preserve model response accuracy while maintaining high hit rates, even without guaranteeing exact routing.

\subsection{CECAR}
\label{sec:cecar}
CECAR is motivated by the observation that MoE models can retain a reasonable level of accuracy even without exact expert routing. This method leverages the output of a learning-based scoring model, and we experimentally confirm that it can achieve higher cache hit rates than LRU. Based on this score, CECAR excludes uncached experts from the router-selected top-$k$ set and reassigns weights to the remaining experts. For example, if the 7th and 8th experts in a top-8 selection are not resident in the cache, normalized weights are reassigned to the remaining top-1 to top-6 experts. In this way, CECAR preserves routing sparsity while maximizing cache hit rates.

Figure~\ref{fig:cecar-routing} illustrates an example of CECAR under a cache size of 24.
In the illustrated top-8 routing decision, three experts are initially not resident in the cache, resulting in an expected cache hit rate of $5/8$.
During expert selection, the bonus mechanism excludes Expert~63 from the routing candidates due to its low expected reuse.
In addition, top-$k$ expert skipping (with $k=4$ in the example) further removes Expert~4 from consideration.
As a result, the effective routing set contains only cached experts, increasing the expected hit-rate from $5/8$ to $6/7$.

As shown in Algorithm \ref{alg:cecar}, CECAR decomposes the routing stage into separate expert selection and weighting steps. A bonus is computed by measuring the margin between the top-1 and top-$k$ router logits, and logit adjustments are applied only to experts that are resident in the cache. The magnitude of this bonus is modulated according to an ML-based eviction score, and does not affect the routing probabilities themselves. The adjusted logits are used exclusively for expert selection (top-$k$), while the final softmax weights are computed from the original logits. This design steers expert selection toward cache-friendly choices while preserving the original weighting behavior of the model.

During the decoding stage, top-$k$ experts are selected using the bonus-adjusted logits; if any of the selected experts are not resident in the cache, they are removed from the selection. Weights are then assigned to the remaining experts by normalizing them with respect to the sum over the original top-8 experts. The top-$k$ threshold used in this process is empirically fixed per model, and is set to 4 for Qwen3-30B-A3B and 3 for DeepSeek-V2-Lite-Chat in our study (see Appendix~\ref{appendix:cecar-topk}). Although this approach introduces a degree of approximation compared to exact top-$k$ routing, it is compensated by token-level soft weighting. As a result, the method achieves high flexibility without incurring performance degradation due to cache misses.

The scoring model used in CECAR leverages the scores produced by the ML-based cache policy predictor described earlier. This score represents the eviction priority of each expert and is directly incorporated into the computation of the logit adjustment coefficient, denoted as $\delta_e$ in Algorithm~\ref{alg:cecar}. Through this design, CECAR enables efficient, scoring-based bonus assignment within the existing MoE architecture without introducing additional parameters.

Experimental results (Table \ref{tab:benchmark}) show that CECAR consistently outperforms MoCCE in terms of both cache hit rate and decode speed. In particular, CECAR consistently maintains an average cache hit rate exceeding 95\% with negligible accuracy degradation in Qwen3-30B-A3B. Furthermore, CECAR is highly practical in that it can be directly applied to existing MoE architectures without requiring any structural modifications.

\begin{algorithm}[t]
\small
\setlength{\algomargin}{8pt}
\caption{CECAR: Cache \& Expert Co-Aware Routing}
\label{alg:cecar}
\KwIn{
Layer index $\ell$, 
expert logits $\{z_e\}_{e=1}^N$,
cache state $\mathcal{C}_\ell$, 
recency scores $r_{\text{norm}}^e$, 
frequency scores $f_{\text{norm}}^e$, 
top-$K$, margin index $m$, prune threshold $\tau$
}

\textbf{(1) Compute routing margin}\\
Let $z_{[1]} \ge z_{[2]} \ge \cdots \ge z_{[N]}$ be the sorted values of $\{z_e\}$, $\Delta \leftarrow z_{[1]} - z_{[m]}$\\

\textbf{(2) Eviction score computation}\\
\ForEach{$e \in \mathcal{C}_\ell$}{
  $s_\ell(e) \leftarrow \mathrm{ML\_Scorer}\big(
    r_{\text{norm}}^e,\,
    f_{\text{norm}}^e
  \big)$
}

\textbf{(3) Eviction-aware cache bonus}\\
\ForEach{$e \in \mathcal{C}_\ell$}{
  $\delta_e \leftarrow 1 - \frac{s_\ell(e)}{\max_{e' \in \mathcal{C}_\ell} s_\ell(e')}$, $z'_e \leftarrow z_e + \Delta \cdot \delta_e$
}

\textbf{(4) Cache-aware top-$k$ selection} $\mathcal{E}_{\text{cand}} \leftarrow \mathrm{TopK}(z', K)$\\

\textbf{(5) Cache-conditioned pruning} $\mathcal{E} \leftarrow \emptyset$\\

\ForEach{$(j,e) \in \mathcal{E}_{\text{cand}}$}{
  \If{$e \in \mathcal{C}_\ell$ \textbf{or} $j \le \tau$}{
    $\mathcal{E} \leftarrow \mathcal{E} \cup \{e\}$
  }
}

\textbf{(6) Cache replacement}\\
\ForEach{$e \in \mathcal{E}$}{
  \If{$e \notin \mathcal{C}_\ell$}{
    $e^{\ast} \leftarrow \arg\max_{e' \in \mathcal{C}_\ell} s_\ell(e')$, $\mathcal{C}_\ell \leftarrow (\mathcal{C}_\ell \setminus \{e^{\ast}\}) \cup \{e\}$
  }
}

\Return{$\mathcal{E}$}
\end{algorithm}

\subsection{ML-Based Cache Policy}
Existing cache policies, such as LRU and LFU, fail to capture the complex expert reuse patterns that arise during MoE inference. FlashMoE addresses this limitation by learning an ML-based eviction policy trained with Belady’s optimal targets, achieving improved hit rates on in-distribution workloads.

However, we observe that FlashMoE’s ML cache policy exhibits two critical limitations when applied in conjunction with expert-aware routing. First, the learned policy fails to generalize to out-of-distribution (OOD) inputs, often underperforming even LRU. Second, its performance is highly sensitive to cache size, degrading significantly under small or varying cache configurations, which limits its applicability in memory-constrained settings such as on-device inference.

To address these issues, we design an ML-based cache policy tailored for CECAR.
Our approach decouples eviction prediction from expert indices, preventing overfitting to dataset-specific expert access patterns observed during training and thereby improving robustness to out-of-distribution inputs.
It produces fine-grained, expert-specific eviction likelihood scores that can be directly incorporated into routing decisions.
Moreover, we account for the fact that expert reuse characteristics vary across cache regimes.
Rather than relying on a single reuse signal, CECAR adjusts the relative influence of multiple reuse predictors according to the cache configuration.
We refer to these predictors as the \emph{hot expert detector} ($\mathrm{FFN}_{\text{Hot}}$) and the \emph{cold expert detector} ($\mathrm{FFN}_{\text{Cold}}$),
which together enable eviction decisions to adapt to different memory constraints.

\begin{figure}[t]
  \centering
  \includegraphics[width=0.66\linewidth]{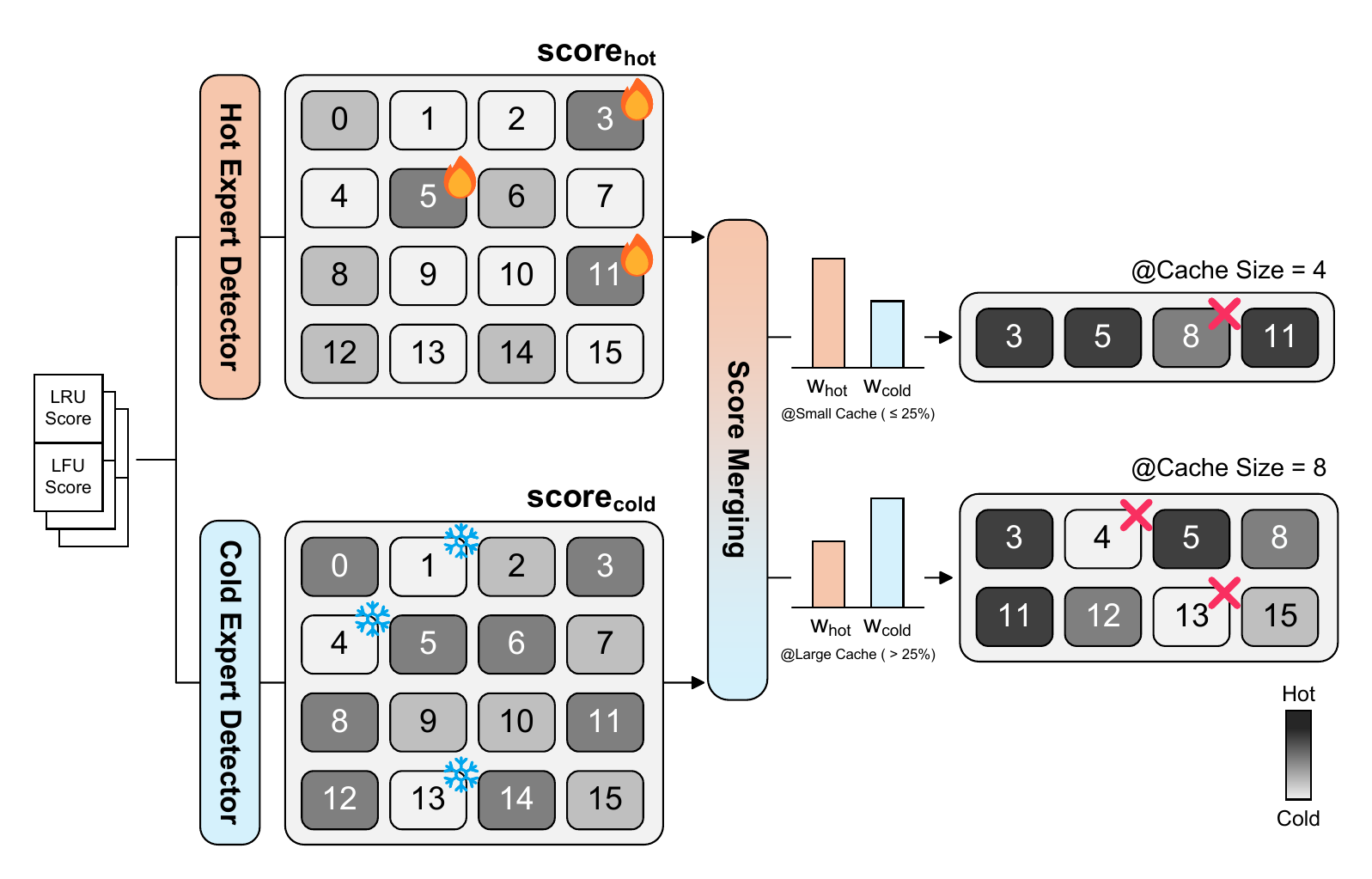}
  \captionsetup{justification=centering}
  \caption{Merging $\mathrm{FFN}_{\text{Hot}}$ and $\mathrm{FFN}_{\text{Cold}}$ for Eviction Score.}
  \label{fig:cecar-eviction}
\end{figure}

Figure~\ref{fig:cecar-eviction} illustrates the principle of hot and cold expert detection used in the proposed ML-based cache policy.
The figure assumes a model with 16 total experts and top-2 routing.
The hot expert detector is trained to identify experts that are likely to be reused in the near term
(e.g., Expert Index 3, 5, and 11),
whereas the cold expert detector is trained to identify experts that are unlikely to be reused
(e.g., Expert Index 1, 4, and 13).
When the cache size is small, caching decisions must carefully distinguish experts that are likely to be reused in the near term
from those that were only recently accessed,
and the hot detector is therefore weighted more heavily.
As the cache size increases, identifying experts with low reuse likelihood becomes more important,
shifting emphasis toward the cold detector.
This cache-size-aware weighting enables robust performance across workloads and cache sizes.

The precise weighting strategy, training targets, and an analysis of the limitations of FlashMoE and how they are addressed by our approach are described in Appendix~\ref{appendix:ml-indetail}.

\section{Experiments}

\subsection{Experiment Setup}
\label{sec:setup}
All experiments are conducted under a controlled inference setting using stochastic decoding with temperature 0.7, top-$p$ sampling threshold of 0.95, and a maximum generation length of 2048 tokens. We evaluate Qwen3-30B-A3B, DeepSeek-V2-Lite-Chat, and OLMoE-1B-7B-0125-Instruct on HumanEval, MBPP\citep{austin2021programsynthesislargelanguage}, GPQA-Diamond, and MATH-500. Accuracy is measured on the full benchmark datasets using a virtual-cache executor of Algorithm~\ref{alg:cecar} with all expert weights resident in VRAM, since routing decisions and cache state depend only on expert identities and not on whether weights are materialized; hit-rate and decode speed are measured on the actual SSD-offloading setup by sampling 100 instances per benchmark (400 samples in total), with host DRAM pinned to $\sim$1~GB (see Appendix~\ref{appendix:measurement}). A fixed cache budget of 24 experts per layer is used across all methods. While the cache size is constant, the specific experts stored are dynamically updated during decoding. Decoding throughput (tokens/s) is measured during the decode phase only. For the hardware configuration of the experiments, please refer to Appendix~\ref{appendix:hardware}.

\subsection{Component-wise Ablation of CECAR}

Table \ref{tab:ablation-ml-bonus-pruning} presents a component-wise ablation study quantifying the effects of the ML-based cache policy, cache-aware bonus routing, and expert pruning on HumanEval accuracy, cache hit rate, and decoding speed. The baseline configuration achieves 47.56 accuracy, a cache hit rate of 54.83\%, and a decoding speed of 3.40 tokens/s.

Introducing the ML-based cache policy alone increases the cache hit rate from 54.83\% to 57.95\% while leaving accuracy unchanged (47.56), resulting in a modest throughput improvement to 3.54 tokens/s. Applying cache-aware bonus routing significantly boosts the cache hit rate to 75.91\% and decoding speed to 5.45 tokens/s; however, this comes at the cost of reduced accuracy (44.51), indicating semantic distortion when routing decisions are biased without eviction awareness.

Combining the ML-based cache policy with bonus routing restores and even improves accuracy to 51.83, while sustaining a high cache hit rate of 76.57\% and decoding speed of 5.60 tokens/s. Similarly, combining the ML-based cache policy with expert pruning improves both accuracy (49.39) and throughput (7.40 tokens/s) compared to pruning alone. When all components are enabled, CECAR achieves the best overall performance, reaching a cache hit rate of 86.87\% and a decoding speed of 8.98 tokens/s, while maintaining strong accuracy (49.39).

These results demonstrate that while bonus routing and expert pruning are effective at improving cache efficiency and throughput, their benefits can be fully realized only when guided by the ML-based cache policy. Learned eviction scores play a critical role in preventing semantic degradation and enabling stable integration of cache-aware routing and pruning under constrained memory budgets.

\begin{table*}[t]
\centering
\scriptsize
\setlength{\algomargin}{8pt}
\caption{Ablation study on ML-based cache policy, CECAR bonus routing, and expert pruning on DeepSeek-V2-Lite-Chat. Accuracy, hit-rate, and decoding speed are all measured on HumanEval only, whereas Table~\ref{tab:benchmark} averages hit-rate and speed over four benchmarks.}
\label{tab:ablation-ml-bonus-pruning}

\begin{tabular}{ccc ccc}
\toprule
\textbf{ML} & \textbf{Bonus} & \textbf{Pruning}
& \textbf{HumanEval} & \textbf{Hit-rate (\%)} & \textbf{Decoding Speed} \\
\midrule
$\times$ & $\times$ & $\times$ & 47.56 & 54.83 & 3.40 \\
\midrule
$\checkmark$ & $\times$ & $\times$ & 47.56 & 57.95 & 3.54 \\
$\times$ & $\checkmark$ & $\times$ & 44.51 & 75.91 & 5.45 \\
$\times$ & $\times$ & $\checkmark$ & 47.56 & 63.47 & 7.35 \\
$\checkmark$ & $\checkmark$ & $\times$ & \textbf{51.83} & 76.57 & 5.60 \\
$\checkmark$ & $\times$ & $\checkmark$ & \underline{49.39} & 66.87 & \underline{7.40} \\
$\times$ & $\checkmark$ & $\checkmark$ & \underline{49.39} & \underline{82.82} & 7.07 \\
$\checkmark$ & $\checkmark$ & $\checkmark$ & \underline{49.39} & \textbf{86.87} & \textbf{8.98} \\
\bottomrule
\end{tabular}
\end{table*}

\begin{table*}[t]
\centering
\small
\caption{Comparison of cache hit rates (\%) across replacement policies.}
\label{tab:cache-policy}

\setlength{\tabcolsep}{4pt}

\resizebox{\linewidth}{!}{
\begin{tabular}{lcccc ccccc}
\toprule
\textbf{Policy}
& \multicolumn{4}{c}{\textbf{Cache Size (TriviaQA)}}
& \multicolumn{5}{c}{\textbf{Dataset Generalization (Cache Size = 24)}} \\
\cmidrule(lr){2-5} \cmidrule(lr){6-10}
& 8 & 16 & 24 & 32
& TriviaQA & HumanEval & MBPP & GPQA-Diamond & MATH-500 \\
\midrule
Belady’s Optimal
& 52.30 & 72.47 & 82.56 & 88.68
& 82.56 & 79.51 & 79.62 & 82.37 & 79.49 \\
\midrule
\textbf{ML (CECAR)}$^\ast$
& \textbf{44.18} & \textbf{62.42} & \textbf{73.63} & \underline{82.04}
& \textbf{73.63} & \textbf{68.37} & \textbf{69.31} & \textbf{73.73} & \textbf{69.63} \\
ML (FlashMoE)
& \underline{34.41} & \underline{58.63} & \underline{73.22} & \textbf{82.23}
& \underline{73.22} & 44.84 & 46.17 & 53.07 & 46.17 \\
LRU
& 33.39 & 58.45 & 69.49 & 78.28
& 69.49 & \underline{65.50} & \underline{65.26} & \underline{70.86} & \underline{66.04} \\
LFU
& 11.33 & 30.65 & 47.53 & 60.96
& 47.53 & 39.50 & 47.25 & 45.66 & 43.49 \\
LIFO
& 9.60 & 27.87 & 43.71 & 56.89
& 43.71 & 39.20 & 45.30 & 39.75 & 39.13 \\
\bottomrule
\end{tabular}
}

\vspace{0.4em}
\footnotesize
$\ast$ All results are obtained with identical Top-$k$ routing; only the cache replacement policy differs.
\end{table*}

\subsection{ML-Based Cache Policy for CECAR}
As shown in Table \ref{tab:cache-policy}, on the TriviaQA\citep{joshi2017triviaqalargescaledistantly} in-distribution (ID) dataset, ML(FlashMoE) achieves higher cache hit rates than LRU and LFU policies. However, on out-of-distribution (OOD) datasets such as HumanEval and MBPP, ML(FlashMoE) exhibits lower hit rates than LRU, indicating limited robustness to workload shifts. In contrast, the proposed ML(CECAR) consistently achieves the highest cache hit rates across all evaluated datasets, demonstrating strong robustness under OOD workloads.

Table \ref{tab:cache-policy} also highlights that the performance of ML (FlashMoE) is highly sensitive to cache size. In particular, when the cache size is small, ML(FlashMoE) underperforms even LRU, including on the TriviaQA in-distribution dataset. By contrast, ML(CECAR) maintains stable and superior hit rates across a wide range of cache sizes, indicating improved robustness to cache capacity variations.

On its own, the ML-based cache policy improves the hit rate over LRU by only 3--4 points (Table~\ref{tab:cache-policy}; 54.83\% to 57.95\% in Table~\ref{tab:ablation-ml-bonus-pruning}). Its value lies in robustness: unlike ML(FlashMoE), it does not fall below LRU on out-of-distribution workloads or at small cache sizes, so its scores can be used in the routing stage without the failure mode seen in Table~\ref{tab:cache-policy}.
This robustness does not come from the scorer's size: a sweep over 15 architectures spanning a $1000\times$ range of parameter counts changes the OOD hit rate by less than 0.5\% (Appendix~\ref{appendix:scorer-sweep}).

\begin{table*}[t]
\centering
\caption{Benchmark accuracy comparison \& performance of cache-aware MoE inference methods. Accuracy is mean $\pm$ SD over 3 decoding seeds; bold and underline mark the best and second-best cache-aware or pruning method (baseline excluded).}
\label{tab:benchmark}
\small

\resizebox{\linewidth}{!}{
\begin{tabular}{ll|cccc|ccc}
\hline
\textbf{Model} & \textbf{Method} 
& HumanEval & MBPP & GPQA-D & MATH-500 
& Avg. top-$k$$^\ast$ & Hit-Rate (\%) & Tokens/s \\
\hline

\multirow{5}{*}{\textbf{Qwen3-A3B}}
& Baseline   & 67.7 {\scriptsize$\pm$1.2} & 64.4 {\scriptsize$\pm$0.9} & 36.7 {\scriptsize$\pm$1.5} & 56.7 {\scriptsize$\pm$2.2} & 8.00 & 68.3 & 2.8 {\scriptsize$\pm$0.31} \\
\hline
& DES        & 65.4 {\scriptsize$\pm$2.1} & 52.1 {\scriptsize$\pm$0.7} & 34.0 {\scriptsize$\pm$0.8} & 38.0 {\scriptsize$\pm$1.2} & 4.67 & 71.3 & 4.7 {\scriptsize$\pm$0.97} \\
& ODP        & 66.3 {\scriptsize$\pm$0.9} & 55.4 {\scriptsize$\pm$1.2} & 31.0 {\scriptsize$\pm$0.8} & 44.1 {\scriptsize$\pm$0.8} & 4.71 & 73.1 & 5.0 {\scriptsize$\pm$1.43} \\
& MoCCE      & \underline{67.5} {\scriptsize$\pm$1.9} & \textbf{61.8} {\scriptsize$\pm$0.5} & \textbf{38.0} {\scriptsize$\pm$2.8} & \underline{55.3} {\scriptsize$\pm$2.8} & 8.00 & \underline{92.3} & \underline{8.6} {\scriptsize$\pm$1.80} \\
& \textbf{CECAR}
              & \textbf{67.9} {\scriptsize$\pm$3.0} & \underline{61.1} {\scriptsize$\pm$1.4} & \underline{36.4} {\scriptsize$\pm$1.8} & \textbf{56.4} {\scriptsize$\pm$1.7} & 7.13 & \textbf{96.6} & \textbf{12.5} {\scriptsize$\pm$1.19} \\
\hline

\multirow{5}{*}{\textbf{DeepSeek-A2.4B}}
& Baseline   & 47.4 {\scriptsize$\pm$0.9} & 45.7 {\scriptsize$\pm$0.4} & 16.8 {\scriptsize$\pm$1.8} & 24.1 {\scriptsize$\pm$0.8} & 6.00 & 58.2 & 4.4 {\scriptsize$\pm$4.87} \\
\hline
& DES        & \underline{45.3} {\scriptsize$\pm$0.9} & \underline{44.6} {\scriptsize$\pm$0.6} & 10.8 {\scriptsize$\pm$0.8} & 22.9 {\scriptsize$\pm$1.4} & 5.07 & 62.2 & 5.0 {\scriptsize$\pm$3.48} \\
& ODP        & 44.5 {\scriptsize$\pm$2.7} & 44.0 {\scriptsize$\pm$0.9} & 11.8 {\scriptsize$\pm$1.5} & 23.1 {\scriptsize$\pm$1.3} & 5.11 & 60.4 & 4.6 {\scriptsize$\pm$2.19} \\
& MoCCE      & 44.9 {\scriptsize$\pm$2.5} & 44.2 {\scriptsize$\pm$0.6} & \underline{16.7} {\scriptsize$\pm$1.3} & \textbf{23.9} {\scriptsize$\pm$2.3} & 6.00 & \underline{83.9} & \underline{7.6} {\scriptsize$\pm$0.55} \\
& \textbf{CECAR}
              & \textbf{45.9} {\scriptsize$\pm$1.3} & \textbf{44.8} {\scriptsize$\pm$1.3} & \textbf{17.5} {\scriptsize$\pm$2.3} & \underline{23.5} {\scriptsize$\pm$1.8} & 5.51 & \textbf{87.1} & \textbf{10.3} {\scriptsize$\pm$3.59} \\
\hline

\multirow{5}{*}{\textbf{OLMoE-A1B}}
& Baseline   & 25.8 {\scriptsize$\pm$1.3} & 26.5 {\scriptsize$\pm$0.6} & 25.9 {\scriptsize$\pm$0.8} & 19.0 {\scriptsize$\pm$0.7} & 8.00 & 74.5 & 8.5 {\scriptsize$\pm$1.97} \\
\hline
& DES        & 21.7 {\scriptsize$\pm$2.7} & 22.7 {\scriptsize$\pm$0.3} & 20.4 {\scriptsize$\pm$0.8} & \underline{17.7} {\scriptsize$\pm$0.3} & 5.79 & 77.6 & 16.2 {\scriptsize$\pm$2.93} \\
& ODP        & 24.6 {\scriptsize$\pm$0.9} & 23.7 {\scriptsize$\pm$0.3} & 21.2 {\scriptsize$\pm$2.2} & \textbf{18.3} {\scriptsize$\pm$0.3} & 5.79 & 78.6 & 18.4 {\scriptsize$\pm$5.69} \\
& MoCCE      & \textbf{25.6} {\scriptsize$\pm$2.4} & \textbf{26.6} {\scriptsize$\pm$0.9} & \underline{26.1} {\scriptsize$\pm$1.1} & 16.7 {\scriptsize$\pm$1.0} & 8.00 & \underline{91.4} & \underline{27.7} {\scriptsize$\pm$6.96} \\
& \textbf{CECAR}
              & \underline{25.2} {\scriptsize$\pm$3.1} & \underline{25.5} {\scriptsize$\pm$1.0} & \textbf{28.8} {\scriptsize$\pm$1.0} & 17.3 {\scriptsize$\pm$0.5} & 7.66 & \textbf{93.5} & \textbf{29.9} {\scriptsize$\pm$4.99} \\
\hline

\end{tabular}
}

\vspace{0.4em}
\footnotesize
$\ast$ Average number of routed experts per layer (excluding shared experts).
Baseline and MoCCE use fixed top-$k$ routing ($k{=}8$ for Qwen3-A3B and OLMoE-A1B, $k{=}6$ for DeepSeek-A2.4B),
while DES, ODP, and CECAR dynamically adjust the effective top-$k$.
Accuracy is reported as mean $\pm$ SD over 3 decoding seeds on the full benchmarks; tokens/s are reported as mean $\pm$ SD across repeated on-device runs (Appendix~\ref{appendix:measurement}).
The single-seed HumanEval sweeps in Appendices~\ref{appendix:benchamrk-ablation}--\ref{appendix:lruvsml} use one of these seeds and therefore differ slightly from the 3-seed means here.
\end{table*}

\subsection{Overall Performance of CECAR}
As shown in Table~\ref{tab:benchmark}, CECAR consistently achieves the highest cache hit rates and decoding throughput across all evaluated models.
For the Qwen3-30B-A3B model, CECAR reaches a cache hit rate of 96.6\% with a decoding speed of 12.5 tokens/s, outperforming both LRU-based routing and prior cache-aware methods. In comparison, MoCCE attains a lower hit rate of 92.3\% and 8.6 tokens/s, while DES and ODP achieve only around 70–73\% hit rates with substantially reduced throughput (4–5 tokens/s).

The speedup is not a result of activating fewer experts. On Qwen3-30B-A3B, DES and ODP activate fewer experts on average (4.67 and 4.71) than CECAR (7.13) but run at 4.7--5.0 tokens/s against CECAR's 12.5. The difference is the hit rate (70--73\% vs.\ 96.6\%). DES and ODP prune by routing weight alone, so many of the experts they keep still have to be read from SSD; CECAR's cache-conditioned pruning drops the low-priority experts that are not resident. Appendix~\ref{appendix:overhead} breaks the per-token latency down and shows that reduced expert loading accounts for nearly all of the gain.

We observe a strong positive correlation between cache hit rate and token decoding speed. Methods with hit rates above 90\% consistently operate at significantly higher throughput, whereas strategies with hit rates around 70\% experience frequent stalls due to expert cache misses. Notably, the throughput improvement achieved by CECAR is disproportionate to its absolute hit-rate gain over LRU-based methods, suggesting that eliminating tail cache misses—rare but expensive expert loads—has a dominant impact on decoding latency.

Combined with the accuracy results, CECAR achieves higher cache hit rates and up to 4.5$\times$ speedup while keeping accuracy close to the exact-routing baseline: over the 12 model--benchmark pairs the mean change is $-0.5$ points, 8 pairs are within 1 point of the baseline, and CECAR is the best or second-best cache-aware or pruning method on 11 of the 12 (Table~\ref{tab:benchmark}).

\section{Conclusion}

We identify expert residency as a key bottleneck in MoE-based LLM inference in on-device environments, and experimentally analyze the necessity of cache-aware routing and cache management to mitigate this issue. Our results show that existing ML-based cache policies fall short of Belady’s optimal, and that this limitation becomes even more pronounced under constrained cache capacity. We further demonstrate that approximate expert selection that accounts for cache locality at the routing stage can simultaneously improve cache hit rates and inference performance while locally constraining semantic distortion. CECAR operates orthogonally to commonly used model compression techniques, such as expert quantization, and can be directly combined with them to further reduce memory requirements. Based on these observations, this paper highlights the need for a new routing paradigm that jointly considers cache and expert reuse state, and argues that the proposed CECAR provides a practical and generalizable solution for on-device MoE inference.

\section*{Impact Statements}
This work improves the efficiency of Mixture-of-Experts (MoE) inference under constrained memory budgets by reducing expert cache misses. The proposed cache-aware routing enables more practical deployment of large MoE models on memory-limited hardware, reducing inference latency. The approach operates purely at the inference and system level and does not modify model architectures. Consequently, it does not introduce new model capabilities or increase risks beyond those inherent to the evaluated models.

\section*{Acknowledgments}
This work was supported by the Institute of Information \& Communications Technology Planning \& Evaluation (IITP) grants funded by the Korean government (MSIT) under Grant Nos.\ RS-2025-02218704 (Development of Chiplet-based DNN-SNN Hybrid Computing Processor for Large-Language-Models, 50\%) and RS-2025-02218733 (Full-Stack Development of a 1-POPS Digital Neuromorphic Event Processor, 50\%).

\bibliographystyle{plainnat}
\bibliography{example_paper,cr_additions}

\begin{thebibliography}{24}
\providecommand{\natexlab}[1]{#1}
\providecommand{\url}[1]{\texttt{#1}}
\expandafter\ifx\csname urlstyle\endcsname\relax
  \providecommand{\doi}[1]{doi: #1}\else
  \providecommand{\doi}{doi: \begingroup \urlstyle{rm}\Url}\fi

\bibitem[Alizadeh et~al.(2024)Alizadeh, Mirzadeh, Belenko, Khatamifard, Cho,
  Del~Mundo, Rastegari, and Farajtabar]{alizadeh-etal-2024-llm}
Keivan Alizadeh, Seyed~Iman Mirzadeh, Dmitry Belenko, S.~Khatamifard, Minsik
  Cho, Carlo~C Del~Mundo, Mohammad Rastegari, and Mehrdad Farajtabar.
\newblock {LLM} in a flash: Efficient large language model inference with
  limited memory.
\newblock In Lun-Wei Ku, Andre Martins, and Vivek Srikumar, editors,
  \emph{Proceedings of the 62nd Annual Meeting of the Association for
  Computational Linguistics (Volume 1: Long Papers)}, pages 12562--12584,
  Bangkok, Thailand, August 2024. Association for Computational Linguistics.
\newblock \doi{10.18653/v1/2024.acl-long.678}.
\newblock URL \url{https://aclanthology.org/2024.acl-long.678/}.

\bibitem[Austin et~al.(2021)Austin, Odena, Nye, Bosma, Michalewski, Dohan,
  Jiang, Cai, Terry, Le, and Sutton]{austin2021programsynthesislargelanguage}
Jacob Austin, Augustus Odena, Maxwell Nye, Maarten Bosma, Henryk Michalewski,
  David Dohan, Ellen Jiang, Carrie Cai, Michael Terry, Quoc Le, and Charles
  Sutton.
\newblock Program synthesis with large language models, 2021.
\newblock URL \url{https://arxiv.org/abs/2108.07732}.

\bibitem[Belady(1966)]{5388441}
L.~A. Belady.
\newblock A study of replacement algorithms for a virtual-storage computer.
\newblock \emph{IBM Systems Journal}, 5\penalty0 (2):\penalty0 78--101, 1966.
\newblock \doi{10.1147/sj.52.0078}.

\bibitem[Chen et~al.(2025{\natexlab{a}})Chen, Xie, Zhang, Tang, Wang, Dong,
  Chen, Yuan, Lin, Qiu, Zhu, Ou, Liao, Chen, Ai, Wu, and
  Zhang]{10.1145/3731569.3764843}
Hongtao Chen, Weiyu Xie, Boxin Zhang, Jingqi Tang, Jiahao Wang, Jianwei Dong,
  Shaoyuan Chen, Ziwei Yuan, Chen Lin, Chengyu Qiu, Yuening Zhu, Qingliang Ou,
  Jiaqi Liao, Xianglin Chen, Zhiyuan Ai, Yongwei Wu, and Mingxing Zhang.
\newblock Ktransformers: Unleashing the full potential of cpu/gpu hybrid
  inference for moe models.
\newblock In \emph{Proceedings of the ACM SIGOPS 31st Symposium on Operating
  Systems Principles}, 2025{\natexlab{a}}.

\bibitem[Chen et~al.(2021)Chen, Tworek, Jun, Yuan, de~Oliveira~Pinto, Kaplan,
  Edwards, Burda, Joseph, Brockman, Ray, Puri, Krueger, Petrov, Khlaaf, Sastry,
  Mishkin, Chan, Gray, Ryder, Pavlov, Power, Kaiser, Bavarian, Winter, Tillet,
  Such, Cummings, Plappert, Chantzis, Barnes, Herbert-Voss, Guss, Nichol,
  Paino, Tezak, Tang, Babuschkin, Balaji, Jain, Saunders, Hesse, Carr, Leike,
  Achiam, Misra, Morikawa, Radford, Knight, Brundage, Murati, Mayer, Welinder,
  McGrew, Amodei, McCandlish, Sutskever, and
  Zaremba]{chen2021evaluatinglargelanguagemodels}
Mark Chen, Jerry Tworek, Heewoo Jun, Qiming Yuan, Henrique~Ponde
  de~Oliveira~Pinto, Jared Kaplan, Harri Edwards, Yuri Burda, Nicholas Joseph,
  Greg Brockman, Alex Ray, Raul Puri, Gretchen Krueger, Michael Petrov, Heidy
  Khlaaf, Girish Sastry, Pamela Mishkin, Brooke Chan, Scott Gray, Nick Ryder,
  Mikhail Pavlov, Alethea Power, Lukasz Kaiser, Mohammad Bavarian, Clemens
  Winter, Philippe Tillet, Felipe~Petroski Such, Dave Cummings, Matthias
  Plappert, Fotios Chantzis, Elizabeth Barnes, Ariel Herbert-Voss,
  William~Hebgen Guss, Alex Nichol, Alex Paino, Nikolas Tezak, Jie Tang, Igor
  Babuschkin, Suchir Balaji, Shantanu Jain, William Saunders, Christopher
  Hesse, Andrew~N. Carr, Jan Leike, Josh Achiam, Vedant Misra, Evan Morikawa,
  Alec Radford, Matthew Knight, Miles Brundage, Mira Murati, Katie Mayer, Peter
  Welinder, Bob McGrew, Dario Amodei, Sam McCandlish, Ilya Sutskever, and
  Wojciech Zaremba.
\newblock Evaluating large language models trained on code, 2021.
\newblock URL \url{https://arxiv.org/abs/2107.03374}.

\bibitem[Chen et~al.(2025{\natexlab{b}})Chen, Wang, Shao, Zeng, Xu, and
  Cheng]{chen2025banpickehancingperformanceefficiency}
Yuanteng Chen, Peisong Wang, Yuantian Shao, Nanxin Zeng, Chang Xu, and Jian
  Cheng.
\newblock Ban\&pick: Ehancing performance and efficiency of moe-llms via smarter
  routing, 2025{\natexlab{b}}.
\newblock URL \url{https://arxiv.org/abs/2509.06346}.

\bibitem[DeepSeek-AI et~al.(2024)DeepSeek-AI, Liu, Feng, Wang, Wang, Liu, Zhao,
  Dengr, Ruan, Dai, Guo, Yang, Chen, Ji, Li, Lin, Luo, Hao, Chen, Li, Zhang,
  Xu, Yang, Zhang, Ding, Xin, Gao, Li, Qu, Cai, Liang, Guo, Ni, Li, Chen, Yuan,
  Qiu, Song, Dong, Gao, Guan, Wang, Zhang, Xu, Xia, Zhao, Zhang, Li, Wang,
  Zhang, Zhang, Tang, Li, Tian, Huang, Wang, Zhang, Zhu, Chen, Du, Chen, Jin,
  Ge, Pan, Xu, Chen, Li, Lu, Zhou, Chen, Wu, Ye, Ma, Wang, Zhou, Yu, Zhou,
  Zheng, Wang, Pei, Yuan, Sun, Xiao, Zeng, An, Liu, Liang, Gao, Zhang, Li, Jin,
  Wang, Bi, Liu, Wang, Shen, Chen, Chen, Nie, Sun, Wang, Liu, Xie, Yu, Song,
  Zhou, Yang, Lu, Su, Wu, Li, Wei, Zhu, Xu, Huang, Li, Zhao, Sun, Li, Wang,
  Zheng, Zhang, Xiong, Zhao, He, Tang, Piao, Dong, Tan, Liu, Wang, Guo, Zhu,
  Wang, Zou, Zha, Ma, Yan, You, Liu, Ren, Ren, Sha, Fu, Huang, Zhang, Xie, Hao,
  Shao, Wen, Xu, Zhang, Li, Wang, Gu, Li, and
  Xie]{deepseekai2024deepseekv2strongeconomicalefficient}
DeepSeek-AI, Aixin Liu, Bei Feng, Bin Wang, Bingxuan Wang, Bo~Liu, Chenggang
  Zhao, Chengqi Dengr, Chong Ruan, Damai Dai, Daya Guo, Dejian Yang, Deli Chen,
  Dongjie Ji, Erhang Li, Fangyun Lin, Fuli Luo, Guangbo Hao, Guanting Chen,
  Guowei Li, H.~Zhang, Hanwei Xu, Hao Yang, Haowei Zhang, Honghui Ding, Huajian
  Xin, Huazuo Gao, Hui Li, Hui Qu, J.~L. Cai, Jian Liang, Jianzhong Guo, Jiaqi
  Ni, Jiashi Li, Jin Chen, Jingyang Yuan, Junjie Qiu, Junxiao Song, Kai Dong,
  Kaige Gao, Kang Guan, Lean Wang, Lecong Zhang, Lei Xu, Leyi Xia, Liang Zhao,
  Liyue Zhang, Meng Li, Miaojun Wang, Mingchuan Zhang, Minghua Zhang, Minghui
  Tang, Mingming Li, Ning Tian, Panpan Huang, Peiyi Wang, Peng Zhang, Qihao
  Zhu, Qinyu Chen, Qiushi Du, R.~J. Chen, R.~L. Jin, Ruiqi Ge, Ruizhe Pan,
  Runxin Xu, Ruyi Chen, S.~S. Li, Shanghao Lu, Shangyan Zhou, Shanhuang Chen,
  Shaoqing Wu, Shengfeng Ye, Shirong Ma, Shiyu Wang, Shuang Zhou, Shuiping Yu,
  Shunfeng Zhou, Size Zheng, T.~Wang, Tian Pei, Tian Yuan, Tianyu Sun, W.~L.
  Xiao, Wangding Zeng, Wei An, Wen Liu, Wenfeng Liang, Wenjun Gao, Wentao
  Zhang, X.~Q. Li, Xiangyue Jin, Xianzu Wang, Xiao Bi, Xiaodong Liu, Xiaohan
  Wang, Xiaojin Shen, Xiaokang Chen, Xiaosha Chen, Xiaotao Nie, Xiaowen Sun,
  Xiaoxiang Wang, Xin Liu, Xin Xie, Xingkai Yu, Xinnan Song, Xinyi Zhou, Xinyu
  Yang, Xuan Lu, Xuecheng Su, Y.~Wu, Y.~K. Li, Y.~X. Wei, Y.~X. Zhu, Yanhong
  Xu, Yanping Huang, Yao Li, Yao Zhao, Yaofeng Sun, Yaohui Li, Yaohui Wang,
  Yi~Zheng, Yichao Zhang, Yiliang Xiong, Yilong Zhao, Ying He, Ying Tang, Yishi
  Piao, Yixin Dong, Yixuan Tan, Yiyuan Liu, Yongji Wang, Yongqiang Guo, Yuchen
  Zhu, Yuduan Wang, Yuheng Zou, Yukun Zha, Yunxian Ma, Yuting Yan, Yuxiang You,
  Yuxuan Liu, Z.~Z. Ren, Zehui Ren, Zhangli Sha, Zhe Fu, Zhen Huang, Zhen
  Zhang, Zhenda Xie, Zhewen Hao, Zhihong Shao, Zhiniu Wen, Zhipeng Xu, Zhongyu
  Zhang, Zhuoshu Li, Zihan Wang, Zihui Gu, Zilin Li, and Ziwei Xie.
\newblock Deepseek-v2: A strong, economical, and efficient mixture-of-experts
  language model, 2024.
\newblock URL \url{https://arxiv.org/abs/2405.04434}.

\bibitem[Eliseev and
  Mazur(2023)]{eliseev2023fastinferencemixtureofexpertslanguage}
Artyom Eliseev and Denis Mazur.
\newblock Fast inference of mixture-of-experts language models with offloading,
  2023.
\newblock URL \url{https://arxiv.org/abs/2312.17238}.

\bibitem[Hendrycks et~al.(2021)Hendrycks, Burns, Kadavath, Arora, Basart, Tang,
  Song, and Steinhardt]{hendrycks2021measuringmathematicalproblemsolving}
Dan Hendrycks, Collin Burns, Saurav Kadavath, Akul Arora, Steven Basart, Eric
  Tang, Dawn Song, and Jacob Steinhardt.
\newblock Measuring mathematical problem solving with the math dataset, 2021.
\newblock URL \url{https://arxiv.org/abs/2103.03874}.

\bibitem[Huang et~al.(2025)Huang, Liao, Liu, He, Tan, Zhang, Li, Liu, and
  QI]{huang2025mixture}
Wei Huang, Yue Liao, Jianhui Liu, Ruifei He, Haoru Tan, Shiming Zhang,
  Hongsheng Li, Si~Liu, and XIAOJUAN QI.
\newblock Mixture compressor for mixture-of-experts {LLM}s gains more.
\newblock In \emph{The Thirteenth International Conference on Learning
  Representations}, 2025.
\newblock URL \url{https://openreview.net/forum?id=hheFYjOsWO}.

\bibitem[Jiang et~al.(2023)Jiang, Sablayrolles, Mensch, Bamford, Chaplot,
  de~las Casas, Bressand, Lengyel, Lample, Saulnier, Lavaud, Lachaux, Stock,
  Scao, Lavril, Wang, Lacroix, and Sayed]{jiang2023mistral7b}
Albert~Q. Jiang, Alexandre Sablayrolles, Arthur Mensch, Chris Bamford,
  Devendra~Singh Chaplot, Diego de~las Casas, Florian Bressand, Gianna Lengyel,
  Guillaume Lample, Lucile Saulnier, Lélio~Renard Lavaud, Marie-Anne Lachaux,
  Pierre Stock, Teven~Le Scao, Thibaut Lavril, Thomas Wang, Timothée Lacroix,
  and William~El Sayed.
\newblock Mistral 7b, 2023.
\newblock URL \url{https://arxiv.org/abs/2310.06825}.

\bibitem[Joshi et~al.(2017)Joshi, Choi, Weld, and
  Zettlemoyer]{joshi2017triviaqalargescaledistantly}
Mandar Joshi, Eunsol Choi, Daniel~S. Weld, and Luke Zettlemoyer.
\newblock Triviaqa: A large scale distantly supervised challenge dataset for
  reading comprehension, 2017.
\newblock URL \url{https://arxiv.org/abs/1705.03551}.

\bibitem[Kamahori et~al.(2024)Kamahori, Gu, Zhu, and
  Kasikci]{kamahori2024fiddler}
Keisuke Kamahori, Yile Gu, Kan Zhu, and Baris Kasikci.
\newblock Fiddler: {CPU}-{GPU} orchestration for fast inference of
  mixture-of-experts models.
\newblock In \emph{5th Workshop on practical ML for limited/low resource
  settings}, 2024.
\newblock URL \url{https://openreview.net/forum?id=WX7lxohjFe}.

\bibitem[Kim et~al.(2026)Kim, Lee, Han, Yoo, and
  Kim]{kim2026flashmoereducingssdio}
Byeongju Kim, Jungwan Lee, Donghyeon Han, Hoi-Jun Yoo, and Sangyeob Kim.
\newblock Flashmoe: Reducing ssd i/o bottlenecks via ml-based cache replacement
  for mixture-of-experts inference on edge devices, 2026.
\newblock URL \url{https://arxiv.org/abs/2601.17063}.

\bibitem[Li et~al.(2024)Li, Wen, Wang, Li, Yuan, Liu, Liu, Xu, Wang, Sun, Kong,
  Wang, Geng, Luan, Jin, Ye, Xiong, Zhang, Li, Xu, Li, Li, Liu, Zhang, and
  Liu]{li2024personalllmagentsinsights}
Yuanchun Li, Hao Wen, Weijun Wang, Xiangyu Li, Yizhen Yuan, Guohong Liu,
  Jiacheng Liu, Wenxing Xu, Xiang Wang, Yi~Sun, Rui Kong, Yile Wang, Hanfei
  Geng, Jian Luan, Xuefeng Jin, Zilong Ye, Guanjing Xiong, Fan Zhang, Xiang Li,
  Mengwei Xu, Zhijun Li, Peng Li, Yang Liu, Ya-Qin Zhang, and Yunxin Liu.
\newblock Personal llm agents: Insights and survey about the capability,
  efficiency and security, 2024.
\newblock URL \url{https://arxiv.org/abs/2401.05459}.

\bibitem[Lu et~al.(2024)Lu, Liu, Xu, Zhou, Huang, Zhang, Yan, and
  Li]{lu-etal-2024-experts}
Xudong Lu, Qi~Liu, Yuhui Xu, Aojun Zhou, Siyuan Huang, Bo~Zhang, Junchi Yan,
  and Hongsheng Li.
\newblock Not all experts are equal: Efficient expert pruning and skipping for
  mixture-of-experts large language models.
\newblock In Lun-Wei Ku, Andre Martins, and Vivek Srikumar, editors,
  \emph{Proceedings of the 62nd Annual Meeting of the Association for
  Computational Linguistics (Volume 1: Long Papers)}, pages 6159--6172,
  Bangkok, Thailand, August 2024. Association for Computational Linguistics.
\newblock \doi{10.18653/v1/2024.acl-long.334}.
\newblock URL \url{https://aclanthology.org/2024.acl-long.334/}.

\bibitem[Muennighoff et~al.(2025)Muennighoff, Soldaini, Groeneveld, Lo,
  Morrison, Min, Shi, Walsh, Tafjord, Lambert, Gu, Arora, Bhagia, Schwenk,
  Wadden, Wettig, Hui, Dettmers, Kiela, Farhadi, Smith, Koh, Singh, and
  Hajishirzi]{muennighoff2025olmoe}
Niklas Muennighoff, Luca Soldaini, Dirk Groeneveld, Kyle Lo, Jacob Morrison,
  Sewon Min, Weijia Shi, Evan~Pete Walsh, Oyvind Tafjord, Nathan Lambert,
  Yuling Gu, Shane Arora, Akshita Bhagia, Dustin Schwenk, David Wadden,
  Alexander Wettig, Binyuan Hui, Tim Dettmers, Douwe Kiela, Ali Farhadi,
  Noah~A. Smith, Pang~Wei Koh, Amanpreet Singh, and Hannaneh Hajishirzi.
\newblock {OLM}oe: Open mixture-of-experts language models.
\newblock In \emph{The Thirteenth International Conference on Learning
  Representations}, 2025.
\newblock URL \url{https://openreview.net/forum?id=xXTkbTBmqq}.

\bibitem[Shazeer et~al.(2017)Shazeer, Mirhoseini, Maziarz, Davis, Le, Hinton,
  and Dean]{shazeer2017}
Noam Shazeer, Azalia Mirhoseini, Krzysztof Maziarz, Andy Davis, Quoc Le,
  Geoffrey Hinton, and Jeff Dean.
\newblock Outrageously large neural networks: The sparsely-gated
  mixture-of-experts layer, 2017.
\newblock URL \url{https://arxiv.org/abs/1701.06538}.

\bibitem[Skliar et~al.(2025)Skliar, van Rozendaal, Lepert, Boinovski, Baalen,
  Nagel, Whatmough, and Bejnordi]{skliar2025mixture}
Andrii Skliar, Ties van Rozendaal, Romain Lepert, Todor Boinovski, Mart~Van
  Baalen, Markus Nagel, Paul~N. Whatmough, and Babak~Ehteshami Bejnordi.
\newblock Mixture of cache-conditional experts for efficient mobile device
  inference.
\newblock \emph{Transactions on Machine Learning Research}, 2025.
\newblock ISSN 2835-8856.
\newblock URL \url{https://openreview.net/forum?id=ul4W26KEKz}.

\bibitem[Tang et~al.(2024)Tang, Liu, Hou, Pu, Wang, Heng, Li, and
  Guo]{tang2024hobbitmixedprecisionexpert}
Peng Tang, Jiacheng Liu, Xiaofeng Hou, Yifei Pu, Jing Wang, Pheng-Ann Heng,
  Chao Li, and Minyi Guo.
\newblock Hobbit: A mixed precision expert offloading system for fast moe
  inference, 2024.
\newblock URL \url{https://arxiv.org/abs/2411.01433}.

\bibitem[Xu et~al.(2024)Xu, Li, Chen, Wang, Gao, Cai, and
  Ling]{xu2024ondevicelanguagemodelscomprehensive}
Jiajun Xu, Zhiyuan Li, Wei Chen, Qun Wang, Xin Gao, Qi~Cai, and Ziyuan Ling.
\newblock On-device language models: A comprehensive review, 2024.
\newblock URL \url{https://arxiv.org/abs/2409.00088}.

\bibitem[Xue et~al.(2025)Xue, Fu, Lu, Mai, and
  Marina]{xue2025moeinfinityefficientmoeinference}
Leyang Xue, Yao Fu, Zhan Lu, Luo Mai, and Mahesh Marina.
\newblock Moe-infinity: Efficient moe inference on personal machines with
  sparsity-aware expert cache, 2025.
\newblock URL \url{https://arxiv.org/abs/2401.14361}.

\bibitem[Yang et~al.(2025)Yang, Li, Yang, Zhang, Hui, Zheng, Yu, Gao, Huang,
  Lv, Zheng, Liu, Zhou, Huang, Hu, Ge, Wei, Lin, Tang, Yang, Tu, Zhang, Yang,
  Yang, Zhou, Zhou, Lin, Dang, Bao, Yang, Yu, Deng, Li, Xue, Li, Zhang, Wang,
  Zhu, Men, Gao, Liu, Luo, Li, Tang, Yin, Ren, Wang, Zhang, Ren, Fan, Su,
  Zhang, Zhang, Wan, Liu, Wang, Cui, Zhang, Zhou, and
  Qiu]{yang2025qwen3technicalreport}
An~Yang, Anfeng Li, Baosong Yang, Beichen Zhang, Binyuan Hui, Bo~Zheng, Bowen
  Yu, Chang Gao, Chengen Huang, Chenxu Lv, Chujie Zheng, Dayiheng Liu, Fan
  Zhou, Fei Huang, Feng Hu, Hao Ge, Haoran Wei, Huan Lin, Jialong Tang, Jian
  Yang, Jianhong Tu, Jianwei Zhang, Jianxin Yang, Jiaxi Yang, Jing Zhou,
  Jingren Zhou, Junyang Lin, Kai Dang, Keqin Bao, Kexin Yang, Le~Yu, Lianghao
  Deng, Mei Li, Mingfeng Xue, Mingze Li, Pei Zhang, Peng Wang, Qin Zhu, Rui
  Men, Ruize Gao, Shixuan Liu, Shuang Luo, Tianhao Li, Tianyi Tang, Wenbiao
  Yin, Xingzhang Ren, Xinyu Wang, Xinyu Zhang, Xuancheng Ren, Yang Fan, Yang
  Su, Yichang Zhang, Yinger Zhang, Yu~Wan, Yuqiong Liu, Zekun Wang, Zeyu Cui,
  Zhenru Zhang, Zhipeng Zhou, and Zihan Qiu.
\newblock Qwen3 technical report, 2025.
\newblock URL \url{https://arxiv.org/abs/2505.09388}.

\bibitem[Yu et~al.(2025)Yu, Cui, Zhang, Wang, and
  Wang]{yu2025taminglatencymemorytradeoffmoebased}
Hanfei Yu, Xingqi Cui, Hong Zhang, Hao Wang, and Hao Wang.
\newblock Taming latency-memory trade-off in moe-based llm serving via
  fine-grained expert offloading, 2025.
\newblock URL \url{https://arxiv.org/abs/2502.05370}.

\end{thebibliography}

\newpage
\appendix

\section{Detailed Analysis of ML-Based Cache Policies}
\label{appendix:ml-indetail}
This appendix provides a detailed analysis of the limitations of existing ML-based cache policies and describes the design of the ML-based cache policy proposed for CECAR.
\subsection{Limitations of Existing ML-Based Cache Policies}
\subsubsection{Limited Out-of-Distribution Robustness}
FlashMoE trains a feed-forward network (FFN) to predict eviction scores using normalized LRU and LFU statistics as inputs. The FFN jointly processes statistics from all experts within a layer and outputs eviction scores for all experts simultaneously. This joint prediction structure introduces a strong coupling between eviction decisions and expert indices. As a result, the model learns dataset-specific expert activation biases during training. When the routing distribution changes under OOD workloads, these learned biases no longer hold, leading to degraded cache performance. Consequently, although FlashMoE improves cache hit rates on the in-distribution TriviaQA dataset, it often underperforms even LRU on OOD datasets such as HumanEval and MBPP.

\subsubsection{Sensitivity to Cache Size}
FlashMoE constructs training targets by approximating Belady’s optimal cache replacement policy. At each timestep, the $N_E$ experts in the layer are ranked according to their reuse distance, and binary labels are assigned by setting the top 75 percent of experts to 1 and the remaining experts to 0. This binarization collapses fine-grained reuse distance information into a coarse eviction signal. Under small cache sizes, where eviction decisions depend on subtle differences in reuse distance, this coarse labeling prevents the model from learning accurate eviction priorities. As a result, FlashMoE exhibits significant performance degradation as cache size decreases, even on in-distribution workloads.

\begin{table}[ht]
\centering
\caption{Training Hyperparameters for ML-Based Cache Policies. Full training details for CECAR are given in Appendix~\ref{appendix:scorer-sweep}.}
\label{tab:ml_cache_hyperparams}

\resizebox{\linewidth}{!}{
\begin{tabular}{lccp{6cm}}
\toprule
\textbf{Hyperparameter} & \textbf{ML Cache (FlashMoE)} & \textbf{ML Cache (CECAR)} & \textbf{Notes} \\
\midrule
\# FFNs per MoE Layer & 1 & 2 & Hot and Cold networks for CECAR \\
\# Layers of FFN & 3 & 3 & Base depth per FFN; sweep over 2--4 in Appendix~\ref{appendix:scorer-sweep} \\
Input Dimension & $2 N_E$ & 2 & 2 features per expert (recency, frequency) \\
Hidden Size & 128 & 32 & Base size; sweep over 8--128 in Appendix~\ref{appendix:scorer-sweep} \\
Output Dimension & $N_E$ & 1 & Eviction score per expert \\
Activation & ReLU & SiLU & Non-linear activation \\
Loss & MSELoss & MSELoss & Regression target \\
Optimizer & AdamW & AdamW & - \\
Learning Rate & 1e-3 & 2e-3 & Constant; no warm-up or decay \\
Weight Decay & - & 0.01 & AdamW default \\
Epochs & - & $\le$200 & Early stopping on training loss, patience 5 \\
$D_\text{cold}$ / $C_\text{cap}$ & - & 18 / 36 & See Appendix~\ref{appendix:scorer-sweep} \\
Training Targets & Binary Belady (top 75\%) & Hot \& Cold targets & Hot = short, Cold = long-term reuse\\
\bottomrule
\end{tabular}
}
\end{table}

\subsection{ML-Based Cache Policy Design for CECAR}

\subsubsection{Expert State Representation (Input Features)}

The eviction scorer in CECAR operates on a compact, expert-level state representation derived from routing history. 
Based on this observation, we represent the state of each expert using two scalar statistics: a recency score $r_t^e$ and a frequency score $f_t^e$.

The recency score represents the number of steps since the expert was last accessed, while the frequency score measures how often the expert has been routed so far. These statistics are updated as follows:
\begin{equation}
\resizebox{0.90\linewidth}{!}{$
r_t^e =
\begin{cases}
1, & \text{if expert } e \text{ is accessed at time } t, \\
r_{t-1}^e + 1, & \text{otherwise},
\end{cases}
\quad
f_t^e =
\begin{cases}
f_{t-1}^e + 1, & \text{if expert } e \text{ is accessed at time } t, \\
f_{t-1}^e, & \text{otherwise}.
\end{cases}
$}
\end{equation}

If an expert has never been accessed, its recency score is initialized to infinity. To ensure stable training and scale-invariant input, both access-recency and access-frequency statistics are normalized before being fed into the FFNs. Experts with smaller recency scores or higher frequency scores are more likely to be routed again in the next steps. The final input to the eviction scorer is constructed for each expert independently, ensuring a fixed input dimension regardless of the number of experts:
\begin{equation}
r_{\text{norm}}^e = \frac{1}{r_t^e}, \quad 
f_{\text{norm}}^e = \frac{f_t^e}{\max_e f_t^e}, \quad
\mathbf{x}_t^e = [r_{\text{norm}}^e \,\|\, f_{\text{norm}}^e].
\end{equation}

More complex representations (e.g., routing-pattern statistics or hidden-state-based features) do not consistently improve eviction decisions and often introduce instability across datasets and cache regimes. In contrast, recency and frequency provide a minimal yet robust signal that generalizes well across models.

However, these signals alone are insufficient to deterministically specify optimal eviction decisions: experts with similar $(r_t^e, f_t^e)$ may exhibit different future reuse depending on layer-specific and contextual factors. This motivates a learned eviction predictor that captures non-linear, context-dependent interactions beyond rule-based policies such as LRU or LFU.

\subsubsection{Expert-Independent FFN Architecture}
To eliminate expert-specific bias, we adopt an expert-independent FFN architecture. Instead of jointly predicting eviction scores for all experts, the model predicts the eviction score of each expert independently using a fixed set of expert-agnostic features. Two FFNs are employed: FFN-Hot for short-term Hot targets and FFN-Cold for long-term Cold targets, capturing fine-grained temporal patterns at different scales. Each network is shared across all experts in a layer, ensuring expert-independent predictions. This design prevents the model from encoding dataset-specific expert indices and enables robust generalization across workloads.

In contrast, prior work such as FlashMoE processes all experts jointly and relies on routing statistics collected from a single dataset (e.g., TriviaQA). As shown in Table~\ref{tab:ml_cache_hyperparams}, the input to the FFNs consists of normalized recency and frequency scores aggregated across all experts. Because the model observes fixed expert indices during training, it can inadvertently learn dataset-specific routing patterns. When evaluated on out-of-distribution workloads, this bias manifests as degraded cache performance, even under otherwise simple policies such as LRU or LFU. By using expert-independent inputs and the FFN-Hot / FFN-Cold design shared across experts, CECAR mitigates this issue and achieves robust generalization across diverse datasets.

\subsubsection{Training Data \& Target Construction}
Routing histories were collected from 512 input sequences sampled from the TriviaQA dataset, each containing 512 tokens; the last 20 sequences are held out and used only for the evaluation in Appendix~\ref{appendix:scorer-sweep}. For every expert and time step, recency and frequency scores were computed as described above, and these statistics were used to calculate the reuse distance of each expert.

From the computed reuse distances, two complementary training targets were derived: a short-term \emph{Hot Target} for fine-grained prioritization among frequently reused experts, and a long-term \emph{Cold Target} for coarse-grained discrimination among rarely reused experts. FFN-Hot and FFN-Cold were trained on their respective targets, capturing caching dynamics across multiple temporal scales while remaining shared across all experts.

\subsubsection{Fine-Grained Training Targets}
To capture expert reuse patterns at multiple temporal scales, we introduce two complementary training targets: a \emph{Hot Target} for short-term reuse and a \emph{Cold Target} for long-term reuse. Two separate networks, FFN-Hot and FFN-Cold, are trained on these targets and shared across all experts in a layer. The Hot and Cold Targets together provide a complete, multi-scale view of eviction likelihood.

\paragraph{Hot Target.}
The \emph{Hot Target} focuses on experts likely to be reused in the near future. 
Let $D_e$ denote the reuse distance of expert $e$, and let $D_\text{hot}$ be the reuse distance of the expert ranked $k$-th in ascending order among all experts. This value is used as the cutoff for constructing the Hot Target.
The Hot Target $T_{\text{hot}}$ is defined as:
\begin{equation}
T_{\text{hot}}(D_e) =
\begin{cases}
\dfrac{D_e - 1}{D_{\text{hot}} - 1}, & \text{if } D_e < D_{\text{hot}}, \\
1, & \text{if } D_e \geq D_{\text{hot}}.
\end{cases}
\label{eq:hot-target}
\end{equation}
This formulation preserves the relative ordering among frequently reused experts while saturating the target for those unlikely to be reused within the cache window.

\paragraph{Cold Target.}
The \emph{Cold Target} captures long-term eviction likelihood that the Hot Target alone cannot distinguish. 
Let $D_\text{cold}$ denote a predefined long-term reuse threshold, chosen such that $D_\text{cold} > D_\text{hot}$. 
The Cold Target $T_\text{cold}$ is defined as:
\begin{equation}
T_{\text{cold}}(D_e) =
\begin{cases}
0, & \text{if } D_e < D_{\text{cold}}, \\
0.5, & \text{if } D_{\text{cold}} \leq D_e < 2 D_{\text{cold}}, \\
1, & \text{if } D_e \geq 2 D_{\text{cold}}.
\end{cases}
\label{eq:cold-target}
\end{equation}

\begin{table}[ht]
\centering
\small
\caption{Example Training Targets for FlashMoE and CECAR.}
\label{tab:example_targets}
\begin{tabular}{c|c|c|c}
\hline
$D_e$ & FlashMoE & \multicolumn{2}{c}{CECAR} \\
\hline
 & Binary Target & Hot Target ($D_\text{hot}=6$) & Cold Target ($D_\text{cold}=15$) \\
\hline
1  & 0 & 0.0 & 0.0 \\
2  & 0 & 0.2 & 0.0 \\
3  & 0 & 0.4 & 0.0 \\
4  & 0 & 0.6 & 0.0 \\
5  & 0 & 0.8 & 0.0 \\
6  & 1 & 1.0 & 0.0 \\
10 & 1 & 1.0 & 0.0 \\
15 & 1 & 1.0 & 0.5 \\
20 & 1 & 1.0 & 1.0 \\
\hline
\end{tabular}
\end{table}

\paragraph{Complementary Nature.}
The Hot Target emphasizes short reuse distances ($D_e < D_{\text{hot}}$) while the Cold Target emphasizes long reuse distances ($D_e \ge D_{\text{cold}}$). By combining these targets, the model captures both short-term and long-term eviction likelihood, allowing fine-grained prioritization for small caches and coarse-grained discrimination for large caches.

\paragraph{Eviction Score Composition.}
The outputs of FFN-Hot and FFN-Cold are combined to produce the final eviction score:

\begin{equation}
s_\ell(e) = \mathrm{ML\_Scorer}\big(
    r_{\text{norm}}^e,\,
    f_{\text{norm}}^e
  \big)
= (1 - \omega) \cdot \text{FFN}_{\text{hot}}\!\left(r_{\text{norm}}^e, f_{\text{norm}}^e\right)
+ \omega \cdot \text{FFN}_{\text{cold}}\!\left(r_{\text{norm}}^e, f_{\text{norm}}^e\right),
\label{eq:score-merge}
\end{equation}
where the weight $\omega \in [0, 0.5]$ is the contribution of the long-term (Cold) signal, and $1-\omega$ that of the short-term (Hot) signal.
Smaller caches prioritize short-term reuse, while larger caches gradually increase the influence of long-term patterns.
Formally, $\omega$ is adjusted based on the cache size $C$ as
\begin{equation}
\omega = \min\Big(0.5,\; \frac{C}{C_\text{cap}} \cdot 0.5\Big),
\label{eq:omega}
\end{equation}
where $C_\text{cap}$ is the cache size at which the Hot and Cold contributions reach a 1:1 ratio.
For $C \to 0$ we have $\omega \to 0$, so the Hot detector dominates; as $C$ grows toward $C_\text{cap}$, long-term eviction signals are incorporated progressively.
The upper bound of $0.5$ enforces the maximum 1:1 weighting between Hot and Cold targets, independent of actual RAM size.
(In the submitted version, the roles of $\omega$ and $1-\omega$ in Eq.~\eqref{eq:score-merge} were printed in the opposite order; the released code implements the form given here.)

\section{Implementation Details of Reproduced Baseline Methods}
\label{appendix:benchamrk-ablation}

To comprehensively evaluate the efficacy of the proposed \textbf{CECAR} framework, we reproduced and analyzed four distinct configurations. These consist of two groups of prior baselines---\textbf{MoCCE} (Baseline 1) and pruning-based methods, \textbf{DES} and \textbf{ODP} (Baseline 2)---alongside \textbf{CECAR-LRU} (Baseline 3), an ablation variant of our method. While the first two baseline groups represent established methodologies in the field, CECAR-LRU is introduced as an ablative benchmark. It enables us to quantify the performance gains specifically attributable to our predictive eviction model by substituting it with a standard Least Recently Used (LRU) policy. All evaluations were conducted within a unified simulation environment to ensure a fair comparison across different cache management and expert pruning strategies.

\subsection{MoCCE}

In reproducing MoCCE, we strictly followed the architectural design proposed in the original work, while performing hyperparameter tuning tailored to our target models: \textbf{Qwen3-30B-A3B} and \textbf{DeepSeek-V2-Lite-Chat}.  
MoCCE introduces a scaling factor $\lambda$ to modulate the degree of cache-awareness during expert routing. We performed a parameter sweep over $\lambda \in \{0.2, 0.4, 0.6, 0.8\}$ using the HumanEval benchmark.

Based on the results, we selected $\lambda = 0.2$, which minimized accuracy degradation. This aligns with the findings in MoCCE~\citep{skliar2025mixture}, where excessive $\lambda$ values lead to accuracy loss due to over-utilization of cached experts.

For fine-grained expert architectures such as Qwen and DeepSeek, we adopted $J=2$ as the default setting, following the recommendations of the original MoCCE study.

\begin{table}[ht]
\centering
\small
\caption{MoCCE parameter sweep on HumanEval.}
\label{tab:mocce_sweep}
\begin{tabular}{c|cc}
\hline
$\lambda$ & Qwen3-30B-A3B & DeepSeek-V2-Lite-Chat \\
\hline
Baseline & 67.68 & 47.56 \\
\hline
0.2 & \textbf{69.51} & \textbf{47.56} \\
0.4 & 64.02 & 46.95 \\
0.6 & 62.19 & 46.34 \\
0.8 & 60.36 & 46.34 \\
\hline
\end{tabular}
\end{table}

\subsection{DES and ODP}

Unlike CECAR, which improves cache locality via routing bonuses, \textbf{DES} and \textbf{ODP} are pruning-based optimizations that reduce computation by skipping experts with negligible contributions.  
We reproduced these methods using the algorithmic framework of MC-MoE~\citep{huang2025mixture}, while adhering to the experimental protocols of Ban\&Pick~\citep{chen2025banpickehancingperformanceefficiency}, which are specifically optimized for fine-grained MoE models such as Qwen and DeepSeek.

\paragraph{DES.}
DES dynamically prunes experts based on routing weight concentration. To establish a common baseline, we first conducted an ablation study on vanilla models by sweeping fixed Top-$k$ expert counts from 1 to 8 without any pruning or cache-aware routing.

\begin{table}[ht]
\centering
\small
\caption{Top-$k$ ablation on HumanEval.}
\label{tab:topk_ablation}
\begin{tabular}{c|c|c|c}
\hline
Top-$k$ & Qwen3-30B-A3B & Top-$k$ & DeepSeek-V2-Lite-Chat \\
\hline
8 & 67.68 & 6 & 48.78 \\
\hline
7 & 68.29 & 5 & 48.78 \\
6 & 65.85 & 4 & 45.12 \\
5 & 64.02 & 3 & 39.63 \\
4 & 57.93 & 2 & 29.88 \\
3 & 36.59 & 1 & 6.71 \\
2 & 22.56 & 0 & 0.00 \\
\hline
\end{tabular}
\end{table}

From this study, we identified $k=5$ as the optimal trade-off point for both architectures. Accordingly, DES was reproduced with a dynamic pruning range of Top-4 to Top-6 experts, ensuring an average activation level consistent with Ban\&Pick.

Layer-wise pruning thresholds $\mu_1^{(l)}$ and $\mu_2^{(l)}$ were computed using medians of routing weight ratios:
\[
r = \frac{w_j}{w_{j+1} + \epsilon}, \quad \epsilon = 10^{-6}.
\]

Calibration was performed using domain-specific datasets: MMLU (Code), MathQA (Math), and ARC-Challenge (General Knowledge), enabling pruning behavior to adapt to domain-dependent expert utilization patterns.

\paragraph{ODP.}
ODP extends DES by incorporating attention-based token significance to prevent pruning of semantically critical tokens. We adopted the Token Significance Index $I_j$ from MC-MoE:
\begin{equation}
I_j = \|\mathbf{t}_j\|_1 \cdot \frac{1}{H} \sum_{h=1}^H \left( \frac{1}{L-j} \sum_{i=j}^{L-1} a_{i,j}^{(h)} \right)
\end{equation}

By calculating this index for each layer during the prefill stage, we identify the top 2\% of tokens as core protection targets. These selected tokens are mandated to activate the maximum expert capacity (Top-6), regardless of whether they satisfy the weight-based pruning condition ($r \geq \mu$). This mechanism effectively preserves essential semantic information and mitigates accuracy degradation caused by the loss of critical experts during the pruning process.

\section{CECAR Top-\texorpdfstring{$k$}{k} Threshold Selection}
\label{appendix:cecar-topk}

CECAR dynamically modulates the number of experts involved in routing by establishing a top-$k$ threshold. While experts within this threshold are always activated to ensure a baseline accuracy floor, any experts exceeding the threshold are conditionally included based on their cache residency. To achieve the most aggressive pruning configuration that does not induce significant accuracy degradation, we configured the thresholds at 4 for Qwen and 3 for DeepSeek. These specific values were identified through a comprehensive parameter sweep as presented in Table \ref{tab:cecar_topk}.

\begin{table}[ht]
\centering
\small
\caption{CECAR Top-$k$ threshold sweep on HumanEval.}
\begin{tabular}{c|c|c|c}
\hline
Top-$k$ & Qwen3-30B-A3B & Top-$k$ & DeepSeek-V2-Lite-Chat \\
\hline
Baseline & 67.68 & Baseline & 47.56 \\
\hline
6 & 68.90 & 4 & 49.39\\
5 & 67.68 & 3 & \textbf{50.61} \\
4 & \textbf{70.73} & 2 & 42.07 \\
3 & 55.49 & 1 & 32.32 \\
\hline
\end{tabular}
\label{tab:cecar_topk}
\end{table}

\section{CECAR-LRU vs.\ CECAR-ML}
\label{appendix:lruvsml}

The impact of ML within the CECAR framework extends far beyond a simple increase in cache hit rates. To isolate the specific contribution of our ML model, we benchmarked it against CECAR LRU, a variant that employs a standard Least Recently Used (LRU) policy to assign cache bonuses to recently activated experts. Across four comprehensive benchmarks, CECAR ML outperformed CECAR LRU. Our comparative analysis demonstrates that the ML approach facilitates more sophisticated cache replacement and routing decisions than conventional heuristic-based methods, thereby driving substantial gains in model performance and accuracy.

\begin{table}[ht]
\centering
\small
\caption{Effect of cache policy on DeepSeek-V2-Lite-Chat.}
\begin{tabular}{c|c|cccc}
\hline
Top-$k$ & Policy & HumanEval & MBPP & GPQA-D & MATH-500 \\
\hline
4 & LRU & 43.90 & 43.40 & \textbf{20.71}& 22.80 \\
  & ML  & \textbf{50.61} & \textbf{44.00} & 17.68 & \textbf{26.60} \\
\hline
3 & LRU & 48.78 & 42.40 & \textbf{20.71}& 26.00 \\
  & ML  & \textbf{49.39} & \textbf{46.20} & 19.70 & \textbf{28.40} \\
\hline
\end{tabular}
\label{tab:cecar_cache_policy}
\end{table}

\section{Routing Behavior Analysis of CECAR}
\label{appendix:routing_analysis}

We analyze how CECAR alters routing behavior relative to the baseline router. Specifically, we examine routing stability, perturbation magnitude, and the preservation of expert specialization.

\subsection{Routing Stability}

We measure routing stability using the average first divergence token (FDT), defined as the generation step at which routing decisions first deviate from the baseline router. 

On Qwen3-30B-A3B (averaged over HumanEval, MBPP, and MATH-500), CECAR achieves a higher FDT (25.4) than MoCCE (20.7) and ODP (9.0), indicating that it preserves the original routing behavior for a longer prefix of generation.

\subsection{Routing Perturbation Magnitude}

We quantify the extent of routing modifications using two metrics: (i) the global influence ratio, i.e., the fraction of routing decisions altered by cache-aware adjustments, and (ii) the normalized weight gap, which measures relative changes in routing weights.

\begin{table}[ht]
\centering
\small
\caption{Routing perturbation analysis.}
\begin{tabular}{l|cc}
\hline
Method & Global Influence Ratio & Normalized Weight Gap \\
\hline
CECAR & \textbf{24.30\%} & \textbf{3.12\%} \\
MoCCE & 29.67\% & 4.73\% \\
\hline
\end{tabular}
\end{table}

CECAR produces fewer routing modifications and smaller weight perturbations than MoCCE, indicating that its adjustments remain localized around the original top-$k$ candidates.

\subsection{Expert Specialization Preservation}

We assess whether CECAR preserves expert specialization by measuring the relative usage of the top experts identified by the baseline router. Across all datasets, the average usage ratio remains above 1 (1.16–1.42), indicating that dominant experts are consistently retained.

Moreover, the identity of top experts varies significantly across datasets, reflecting domain-specific specialization. CECAR closely follows these patterns, suggesting that it preserves both expert diversity and domain-adaptive routing behavior.

CECAR improves performance while maintaining routing stability and preserving expert specialization. This indicates that its gains arise from controlled, localized adjustments rather than large deviations from the original routing decisions.

\section{Hardware Configuration of CECAR}
\label{appendix:hardware}

Table~\ref{table:spec} summarizes the hardware and software configuration used for evaluating CECAR. All experiments were conducted on a consumer-grade desktop system with 64~GB of system DRAM and a 16~GB GPU. This configuration was chosen to allow efficient experimentation while retaining practical relevance to on-device deployment scenarios with limited GPU and system memory.  

{\small
\begin{table}[ht]
  \centering
  \caption{Specifications of Desktop System used for CECAR. The system includes ample DRAM to facilitate experimentation, but during inference only a small portion is used to emulate memory-constrained edge environments.}
  \label{table:spec}
  \resizebox{\linewidth}{!}{
  \begin{tabular}{c|c}
    \hline
    \textbf{Component} & \textbf{Specification} \\
    \hline
    CPU & AMD Ryzen 5 9600X (Zen 5, 6 cores, PCIe 5.0 support) \\
    \hline
    GPU & NVIDIA RTX 5070 Ti (16~GB GDDR7, PCIe 5.0) \\
    \hline
    Motherboard & Gigabyte X870 AORUS EAGLE WiFi7 (PCIe 5.0) \\
    \hline
    SSD & SK hynix Platinum P51 NVMe SSD (PCIe 5.0, 7.4~GB/s sequential read) \\
    \hline
    RAM & 16~GB DDR5-6000MHz $\times$ 4 (Dual Channel, 64~GB total installed) \\
    \hline
    OS & Ubuntu 24.04 LTS \\
    \hline
    Environment & Python 3.11, PyTorch 2.7, CUDA 12.9 \\
    \hline
    DRAM Usage During Inference & $\sim 1$~GB reserved for OS and bookkeeping;\\
    \hline
    Expert Storage & Non-cached experts fetched from SSD to emulate edge-device constraints\\
    \hline
  \end{tabular}
  }
\end{table}
}

\subsection*{Emulating Memory-Constrained On-Device Environments}
Although the host machine provides ample DRAM (64~GB), real-world edge devices such as laptops, tablets, or mobile phones typically have far less memory available for model inference. To realistically simulate such constraints, we deliberately restrict the usage of host DRAM for expert parameters during inference. Specifically, we cap system DRAM at 1GB for operating system stability and runtime bookkeeping. The remaining DRAM is not used for storing expert weights.  

Under this configuration, experts that are not currently resident in the cache are fetched directly from SSD storage. This setup emulates the slower storage tier typical in edge devices, where it is not feasible to keep all expert parameters in RAM. By explicitly controlling DRAM usage in this way, we ensure that the observed performance—such as cache hit rates, throughput, and latency—reflects on-device memory limitations rather than the full capacity of the desktop host system.  

Concretely, for Qwen3-30B-A3B, each expert contains approximately 4.7M parameters, corresponding to 9.44~MB under BF16 precision.
Storing all experts therefore requires approximately 56.7~GB of memory, making it infeasible to keep expert parameters in host DRAM under realistic on-device constraints. Accordingly, in our setting with a cache size of 24, expert parameters occupy approximately 10.6~GB of the 16~GB GPU VRAM, while the remaining VRAM is used to store non-expert model parameters and the KV cache.
As a result, CECAR operates with only approximately 1~GB of host DRAM reserved for operating system stability and runtime bookkeeping, whereas a baseline configuration would require 64~GB of host DRAM to keep all experts resident.

In short, although the experiments run on a desktop with sufficient memory, the \textbf{behavior of the system is constrained to mirror realistic edge deployment conditions}. This design allows us to evaluate CECAR’s performance under practical memory limitations without requiring access to specialized hardware with limited RAM.  

\subsection*{Measurement Methodology and Reproducibility}
\label{appendix:measurement}

\paragraph{Two measurement paths.}
The numbers in Table~\ref{tab:benchmark} come from two programs that run the same rule (Algorithm~\ref{alg:cecar}). Accuracy, hit rate, and average top-$k$ come from the evaluation path in the released code (\texttt{./evaluation -{}-bonus\_strategy cecar}). It keeps every expert in VRAM on a separate large-memory GPU, tracks the cache as a set of expert identities, applies the same selection, top-$K$, cache-conditioned pruning (rank $\le \tau$ or resident, no refill), and admit/evict steps as the offloading engine, and runs the full MoE forward pass. Whether an expert is a hit or a miss depends only on identities and cache state, not on where the weights are stored, so this path reports the same hit rate and average top-$k$ as the offloading engine while running fast enough to cover the full benchmarks. Decode speed is the one quantity that needs wall-clock timing on the offloading engine. It is measured on the testbed of Table~\ref{table:spec} with host DRAM limited to about 1~GB with \texttt{mlock()}, on 100 prompts per benchmark. Without the lock, the OS uses free host memory as a page cache, the experts end up resident in DRAM, and no SSD read takes place. Table~\ref{tab:ablation-ml-bonus-pruning} and Table~\ref{tab:cecar_topk} use the same two paths, restricted to HumanEval.

\paragraph{Run-to-run variance.}
Expert selection in Algorithm~\ref{alg:cecar} is deterministic for a given prompt and cache state, but wall-clock speed varies between runs because of token sampling and I/O timing. Table~\ref{tab:speed-variance} gives the decode speed of each method as mean $\pm$ SD over repeated runs of the 100-prompt measurement; Table~\ref{tab:benchmark} reports the same values.

\begin{table}[ht]
\centering
\small
\caption{Decode speed (tokens/s), mean $\pm$ SD across repeated on-device runs (cache size 24 experts/layer, 100 prompts per benchmark).}
\label{tab:speed-variance}
\begin{tabular}{l|ccc}
\toprule
Strategy & Qwen3-A3B & DeepSeek-A2.4B & OLMoE-A1B \\
\midrule
Baseline & 2.8 $\pm$ 0.31  & 4.4 $\pm$ 4.87  & 8.5 $\pm$ 1.97 \\
DES      & 4.7 $\pm$ 0.97  & 5.0 $\pm$ 3.48  & 16.2 $\pm$ 2.93 \\
ODP      & 5.0 $\pm$ 1.43  & 4.6 $\pm$ 2.19  & 18.4 $\pm$ 5.69 \\
MoCCE    & 8.6 $\pm$ 1.80  & 7.6 $\pm$ 0.55  & 27.7 $\pm$ 6.96 \\
\textbf{CECAR} & \textbf{12.5 $\pm$ 1.19} & \textbf{10.3 $\pm$ 3.59} & \textbf{29.9 $\pm$ 4.99} \\
\bottomrule
\end{tabular}
\end{table}

\paragraph{Released code.}
The repository (\url{https://github.com/hoonki2002/CECAR}) contains the evaluation path, the SSD-offloading engine, the throughput harness used for the 100-prompt measurements, and the \texttt{mlock()} script that limits host DRAM. Its default decoding settings are those of Section~\ref{sec:setup} (temperature 0.7, top-$p$ 0.95). An earlier version of the engine refilled the pruned expert set to the full budget after the keep test; the released engine follows the no-refill rule of Algorithm~\ref{alg:cecar}, as the evaluation path does.

\subsection*{Robustness Across Hardware Interfaces}

We evaluate CECAR under different interconnect configurations (PCIe 5.0 vs. PCIe 4.0) to assess hardware sensitivity. Table~\ref{table:pcie} reports expert loading latency and effective throughput.

{\small
\begin{table}[ht]
\centering
\caption{Expert loading latency and throughput across PCIe generations.}
\label{table:pcie}
\resizebox{\linewidth}{!}{
\begin{tabular}{l|c|c|c|c|c}
\toprule
Model & Size & PCIe 5.0 Latency (ms) & PCIe 4.0 Latency (ms) & PCIe 5.0 TP (GB/s) & PCIe 4.0 TP (GB/s) \\
\midrule
OLMoE-1B-7B-0125-Instruct & 12.6M & 5.137 & 6.256 & 2.45 & 2.01 \\
DeepSeek-V2-Lite-Chat     & 17.3M & 7.111 & 8.187 & 2.43 & 2.11 \\
Qwen3-30B-A3B             & 9.4M  & 4.464 & 4.424 & 2.12 & 2.13 \\
\bottomrule
\end{tabular}
}
\end{table}
}

Across all models, the latency difference between PCIe 5.0 and PCIe 4.0 is limited (0.99×–1.22×), indicating weak sensitivity to interconnect bandwidth. This is because individual expert parameters are relatively small (e.g., $\sim$9.4~MB for Qwen3-30B-A3B), which prevents saturation of peak PCIe bandwidth.

Consequently, expert loading latency is dominated by per-transfer overhead rather than bandwidth, and the relative performance of CECAR remains stable across PCIe generations. This suggests that CECAR's gains are primarily driven by routing and cache decisions, rather than hardware-specific bandwidth advantages, and thus generalize across commodity systems. This further indicates that improving expert locality is more critical than increasing interconnect bandwidth in on-device MoE inference.

\section{Strategy Predictions on Sample Problems}

For illustration purposes, we include example problems to demonstrate how different caching strategies generate outputs. 
Each example shows, for each strategy (Baseline, DES, ODP, MoCCE, and CECAR), how the generated tokens differ in both content and quantity. 
These results were obtained by running two representative Math problems through DeepSeek-V2-Lite-Chat, highlighting the differences in the generated tokens produced by each strategy.

\begin{table}[ht]
\small
\centering
\caption{Comparison of strategy outputs on Examples.}

\begin{tabular}{l|c|c|c|c|c}
\hline
Strategy & Example 1 & Example 2 & Avg. top-$k$$^\ast$ & Hit-Rate (\%) & Tokens/s \\
\hline
Baseline & False & True  & 6.00 & 58.78 & 3.24 \\
DES      & False & False & 4.98 & 57.08 & 3.74 \\
ODP      & False & False & 5.08 & 56.95 & 3.79 \\
MoCCE    & True  & False & 6.00 & 87.54 & 8.78 \\
CECAR    & True  & True  & 5.48 & 88.47 & 10.12 \\
\hline
\end{tabular}

\vspace{0.4em}
\footnotesize
$\ast$ Average number of routed experts per layer, excluding shared experts.

\end{table}

\subsection{Example 1}
\textit{\textbf{Question:} If $f(x) = (3x-2)/(x-2)$, what is the value of $f(-2) + f(-1) + f(0)$? Please reason step by step and put your final answer within [answer].}

\textbf{Baseline \textcolor{red}{(False)} :} To find the value of $f(-2) + f(-1) + f(0)$, we need to substitute $x = -2$, $x = -1$, and $x = 0$ into the function $f(x)$ and then add the results. First, let's find $f(-2)$: $f(-2) = (3(-2) - 2)/(-2 - 2) = (-6 - 2)/(-4) = -8/-4 = 2$. Next, let's find $f(-1)$: $f(-1) = (3(-1) - 2)/(-1 - 2) = (-3 - 2)/(-3) = -5/-3 = 5/3$. Finally, let's find $f(0)$: $f(0) = (3(0) - 2)/(0 - 2) = -2/-2 = 1$. Now, we add these results together: $f(-2) + f(-1) + f(0) = 2 + 5/3 + 1  \textcolor{red!75}{= 6/3 + 5/3 + 1/3} = 12/3 = [4]$.

\textbf{DES \textcolor{red}{(False)}:} It seems like you're asking for the value of a function $f(x)$ at two specific points, and then adding those values together. However, the function $f(x)$ is defined as a fraction, and \textcolor{red!75}{we can't simply substitute the values into the fraction to get the result}. The function $f(x)$ is defined as a fraction, and we can't simply substitute the values into the fraction to get the result.

\textbf{ODP \textcolor{red}{(False)}:} To solve this problem, we need to \textcolor{red!75}{find the value of $f(-2)$ and $f(0)$} and then substitute them into the equation. Step 1: Find $f(-2)$. We have $f(x) = (3x-2)/(x-2)$. Substituting $x = -2$ into the equation, we get $f(-2) = (3(-2)-2)/(-2-2) = -8/-4 = 2$. Step 2: Find $f(0)$. We have $f(x) = (3x-2)/(x-2)$. Substituting $x = 0$ into the equation, we get $f(0) = (3(0)-2)/(0-2) = -2/-2 = 1$. Step 3: Substitute $f(-2)$ and $f(0)$ into the equation. We have $f(-2) + f(0) = 2 + 1 = 3$. Therefore, the answer is $[3]$.

\textbf{MoCCE \textcolor{blue}{(True)}:} To find the value of $f(-2) + f(-1) + f(0)$, we need to substitute $x = -2$, $x = -1$, and $x = 0$ into the function $f(x)$. First, let's find $f(-2)$: $f(-2) = (3(-2)-2)/(-2-2) = (-6-2)/(-4) = -8/-4 = 2$. Next, let's find $f(-1)$: $f(-1) = (3(-1)-2)/(-1-2) = (-3-2)/(-3) = -5/-3 = 5/3$. Finally, let's find $f(0)$: $f(0) = (3(0)-2)/(0-2) = -2/-2 = 1$. Now, we add these values together: $f(-2) + f(-1) + f(0) = 2 + 5/3 + 1 = 2 + 5/3 + 3/3 = 2 + 8/3 = 6/3 + 8/3 = 14/3$. So, the value of $f(-2) + f(-1) + f(0)$ is $[14/3]$.

\textbf{CECAR \textcolor{blue}{(True)}:} To find the value of $f(-2) + f(-1) + f(0)$, we need to substitute $x = -2$, $x = -1$, and $x = 0$ into the function $f(x)$. First, let's find $f(-2)$: $f(-2) = (3(-2)-2)/(-2-2) = (-6-2)/(-4) = -8/-4 = 2$. Next, let's find $f(-1)$: $f(-1) = (3(-1)-2)/(-1-2) = (-3-2)/(-3) = -5/-3 = 5/3$. Finally, let's find $f(0)$: $f(0) = (3(0)-2)/(0-2) = -2/-2 = 1$. Now, we can add these values together: $f(-2) + f(-1) + f(0) = 2 + 5/3 + 1 = 2 + 5/3 + 3/3 = 2 + 8/3 = 6/3 + 8/3 = 14/3$. So, the value of $f(-2) + f(-1) + f(0)$ is $[14/3]$.

\subsection{Example 2}
\textit{\textbf{Question:} When the positive integers with exactly three positive divisors are listed in ascending order, what is the fifth number listed? Please reason step by step and put your final answer within [answer].}

\noindent
\textbf{Baseline \textcolor{blue}{(True)} :} The positive integers with exactly three positive divisors are the squares of prime numbers. This is because the divisors of such a number are the number 1, the prime number itself. The first four squares of prime numbers are 4, 9, 25, and 49. Therefore, the fifth number with exactly three positive divisors is [121].

\noindent
\textbf{DES \textcolor{red}{(False)} :} A positive integer with exactly three positive divisors \textcolor{red!75}{must be prime}. This is because a prime number has exactly two positive divisors: 1 and the number itself. So, we need to find the fifth prime number. The first four prime numbers are 2, 3, 5, and 7. Therefore, the fifth prime number is 11. So, the fifth number listed is [11].

\noindent
\textbf{ODP \textcolor{red}{(False)} :} A positive integer with exactly three positive divisors \textcolor{red!75}{must be in the form $p^2$ or $p^3$, where $p$ is prime}. We can list the numbers in ascending order as follows: 
$4 = 2^2$, $9 = 3^2$, $8 = 2^3$, $27 = 3^3$, $16 = 2^4$, $81 = 3^4$, $32 = 2^5$, $243 = 3^5$, $64 = 2^6$, $729 = 3^6$, $128 = 2^7$, $2187 = 3^7$, $256 = 2^8$, $4993 = 3^8$, $512 = 2^9$, $10000 = 2^{10}$, $29169 = 3^9$, $1024 = 2^{11}$, $20000 = 2^{12}$, $30576 = 3^{10}$, $104976 = 2^{13}$, $205376 = 2^{14}$, $311731 = 3^{11}$, $1073744 = 2^{15}$, $2111088 = 2^{16}$, $3229225 = 3^{12}$, $10995112 = 2^{17}$, $21990240 = 2^{18}$, $33554433 = 3^{13}$, $112589994 = 2^{19}$, ...

\noindent
\textbf{MoCCE \textcolor{red}{(False)} :} The positive integers with exactly three positive divisors are \textcolor{red!75}{the square numbers}. This is because the divisors of a square number are 1, the number itself, and the square root of the number. Since the square root must be a positive integer, there can be at most three divisors. The first few square numbers are $1^2 = 1$, $2^2 = 4$, $3^2 = 9$, $4^2 = 16$, $5^2 = 25$, $6^2 = 36$, $7^2 = 49$, $8^2 = 64$, $9^2 = 81$, $10^2 = 100$, and so on. We can see that the fifth square number is $5^2 = 25$. Therefore, the fifth number listed is [25].

\noindent
\textbf{CECAR \textcolor{blue}{(True)} :} The positive integers with exactly three positive divisors are the squares of prime numbers. This is because the divisors of a number are determined by its prime factorization. If a number has exactly three divisors, it must be the square of a prime number, because the square of a prime number has exactly three divisors: 1, the prime number itself, and the square of the prime number. The first four squares of prime numbers are $2^2 = 4$, $3^2 = 9$, $5^2 = 25$, and $7^2 = 49$. The fifth square of a prime number is $11^2 = 121$. Therefore, the fifth number listed when the positive integers with exactly three positive divisors are listed in ascending order is [121].

\section{Comparison with MoE-Infinity}
\label{appendix:offloading-comparison}

MoE-Infinity\citep{xue2025moeinfinityefficientmoeinference} is open-source and targets the same single-request, personal-machine setting as CECAR. We replayed its cache logic (\texttt{ExpertTracer}, \texttt{find\_most\_similar}) on routing traces under the same conditions as our experiments (Qwen3-30B-A3B, 24 experts per layer, top-$k$ 8) and compared the resulting hit rate with LRU and with Belady's optimal policy computed on the original routing. Belady on the original routing is an upper bound for any system that keeps exact routing and only decides which experts to cache or prefetch.

\begin{table}[ht]
\centering
\small
\caption{Hit rate on the four evaluation benchmarks (Qwen3-30B-A3B, 24 experts/layer), mean $\pm$ std over sequences.}
\label{tab:moe-infinity}
\resizebox{\linewidth}{!}{
\begin{tabular}{l|cccc|c}
\toprule
Dataset & MoE-Infinity & LRU & Belady (original routing) & CECAR & $n$ (trace / CECAR) \\
\midrule
MBPP         & 0.516 $\pm$ 0.028 & 0.642 $\pm$ 0.021 & 0.797 $\pm$ 0.013 & \textbf{0.950 $\pm$ 0.019} & 50 / 100 \\
HumanEval    & 0.670 $\pm$ 0.026 & 0.627 $\pm$ 0.027 & 0.731 $\pm$ 0.016 & \textbf{0.949 $\pm$ 0.021} & 6$^\dagger$ / 100 \\
MATH-500     & 0.598 $\pm$ 0.042 & 0.629 $\pm$ 0.017 & 0.718 $\pm$ 0.013 & \textbf{0.968 $\pm$ 0.014} & 6$^\dagger$ / 100 \\
GPQA-Diamond & 0.604 $\pm$ 0.030 & 0.646 $\pm$ 0.033 & 0.737 $\pm$ 0.016 & \textbf{0.982 $\pm$ 0.011} & 6$^\dagger$ / 100 \\
\bottomrule
\end{tabular}
}

\vspace{0.4em}
\footnotesize
$\dagger$ The MBPP trace has 50 sequences. The HumanEval, MATH-500, and GPQA-Diamond traces have 6 sequences with 32 decode tokens each and should be read as indicative only. CECAR's numbers come from its own 100-sample run, not from the same trace.
\end{table}

CECAR's hit rate (0.95--0.98) is above the Belady bound on the original routing (0.72--0.80). A caching or prefetching system cannot pass this bound without changing the routing; CECAR passes it because it changes the routing (about 28\% of selections differ from the baseline router). The two kinds of system therefore act at different points of the pipeline and can be combined: CECAR reduces the number of non-resident experts that are requested, and a prefetcher can hide the latency of the ones that remain.

We could not compare end-to-end throughput. MoE-Infinity allocates every expert in host DRAM (\texttt{AllocateMemory} in \texttt{model\_topology.cpp}) and has no budget or eviction mechanism, so it needs host memory on the order of the full model, about 57~GB for Qwen3-30B-A3B. It ran out of memory when loading this model with 60~GB of RAM, in two attempts. CECAR runs with about 1~GB of host DRAM and reads from SSD on every token, so there is no host-memory budget under which both systems run.

\section{Runtime Overhead Analysis}
\label{appendix:overhead}

Table~\ref{tab:overhead-breakdown} breaks the per-token decode latency of Qwen3-30B-A3B (24 experts per layer) into expert loading, fixed compute, and the cost CECAR adds. The inputs are per-operation costs measured on the testbed of Table~\ref{table:spec}: 2.62~ms per expert miss, 384 expert activations per token (48 layers $\times$ 8 experts), and 42.73~ms of compute per token that does not depend on the cache policy (expert compute 18.61~ms, attention 12.13~ms, norm and gate 11.99~ms).

\begin{table}[ht]
\centering
\small
\caption{Per-token latency at three operating points (Qwen3-30B-A3B, 24 experts/layer).}
\label{tab:overhead-breakdown}
\resizebox{\linewidth}{!}{
\begin{tabular}{l|c|c|c|c|c|c}
\toprule
Operating point & Misses & Expert load (ms) & Fixed compute (ms) & CECAR scorer (ms) & Total (ms) & Tokens/s \\
\midrule
On-demand (0\% hit)      & 384 & 1006.1 & 42.73 & --   & 1048.8 & 0.95 \\
LRU (64.4\% hit, MBPP)   & 137 & 358.2  & 42.73 & --   & 400.9  & 2.49 \\
CECAR (96.6\% hit)       & 13  & 34.2   & 42.73 & 5.27 & 82.2   & 12.16 \\
\bottomrule
\end{tabular}
}
\end{table}

The model gives 12.16 tokens/s at CECAR's operating point, close to the 12.5 tokens/s measured in Table~\ref{tab:benchmark}. Expert loading is the only term that depends on the policy. Going from on-demand loading to CECAR removes 972~ms of loading per token. What CECAR adds is small: the cache-aware selection and logit adjustment cost a few microseconds per layer, the same order as the gate itself, and the eviction scorer costs 5.27~ms per token in the per-layer implementation used for the results in this paper (48 kernel launches per token, dominated by launch overhead) or 0.088~ms when the 48 layers are scored in one batched launch. In either case the added compute is under 0.6\% of the loading it saves. The 2.62~ms miss cost is calibrated against the SSD; if experts were served from the OS page cache instead, the true SSD cost would be higher and the share of the speedup due to loading would be larger still.

\section{Robustness to Cache Budget}
\label{appendix:dynamic-budget}

\subsection{Static Capacity Sweep}
Table~\ref{tab:cache-policy} varies the cache size but reports only hit rate on TriviaQA. On DeepSeek-V2-Lite-Chat, HumanEval accuracy under CECAR is 43.90 at $C{=}8$, 45.73 at $C{=}16$, 49.39 at $C{=}24$, and 50.00 at $C{=}32$. Accuracy falls gradually as the budget shrinks and is flat above $C{=}24$.

\subsection{Budgets That Change During Decoding}
The memory available on a device is not constant, and $\omega(C)$ in Eq.~\eqref{eq:omega} is defined for a changing cache size. We ran three schedules in which the per-layer budget $C_t$ changes during generation, on DeepSeek-V2-Lite-Chat over all 164 HumanEval problems. When the budget shrinks, the excess experts are evicted at once in order of eviction score; $\omega$ is recomputed from $C_t$ at every decode step.
\begin{itemize}
  \item \textbf{S1}: $24 \rightarrow 8$ during decode steps 192--384, then back to 24.
  \item \textbf{S2}: $32 \rightarrow 16$ at step 256.
  \item \textbf{S3}: alternating between 24 and 12 every 64 steps.
\end{itemize}

\begin{table}[ht]
\centering
\small
\caption{Accuracy and hit rate under changing budgets, with a fixed budget of the same time-averaged size for reference (DeepSeek-V2-Lite-Chat, HumanEval, 164 problems).}
\label{tab:dynamic-budget}
\resizebox{\linewidth}{!}{
\begin{tabular}{l|cc|l}
\toprule
Schedule & pass@1 & Hit rate & Fixed-budget reference \\
\midrule
S1 ($24 \rightarrow 8 \rightarrow 24$) & 72/164 & 85.4\% & $C{=}23$ (realized average 22.8): 73/164, 86.5\% \\
S2 ($32 \rightarrow 16$)               & 71/164 & 90.4\% & $C{=}32$ (realized average 31.7$^\dagger$): 71/164, 90.7\% \\
S3 ($24 \leftrightarrow 12$)           & 72/164 & 82.1\% & $C{=}18$ (realized average 20.6): 70/164, 81.4\% \\
\bottomrule
\end{tabular}
}

\vspace{0.4em}
\footnotesize
$\dagger$ Most generations end before step 256, so the realized average budget of S2 is 31.7; we compare against the nearest fixed budget.
\end{table}

\begin{table}[ht]
\centering
\small
\caption{Hit rate immediately after each budget change, compared with the steady-state hit rate of the new budget.}
\label{tab:dynamic-transient}
\begin{tabular}{l|c|l}
\toprule
Transition & After transition & Steady state at new $C$ \\
\midrule
S1: $24 \rightarrow 8$   & 59.2\% & 56.9\% \\
S2: $32 \rightarrow 16$  & 81.4\% & 78.3\% \\
S1: $8 \rightarrow 24$   & 80.2\% & 87.2\% (reached as the cache refills) \\
\bottomrule
\end{tabular}
\end{table}

Under all three schedules, accuracy and hit rate are within one problem and about 1~pp of a fixed budget with the same time-averaged size (Table~\ref{tab:dynamic-budget}). Right after a shrink, the hit rate is already at or above the steady-state hit rate of the smaller budget (Table~\ref{tab:dynamic-transient}), because the experts that survive the shrink are the ones with the best eviction scores. Freezing $\omega$ at its $C{=}24$ value throughout S1 changes the hit rate by at most 0.3~pp and the accuracy by no problems, so the adaptation comes from the scorer rather than from the $\omega$ schedule.

\section{Eviction-Scorer Training and Architecture Sweep}
\label{appendix:scorer-sweep}

$\mathrm{FFN}_{\text{Hot}}$ and $\mathrm{FFN}_{\text{Cold}}$ are per-layer MLPs trained jointly on the summed MSE loss with AdamW, one 512-token sequence per step. Traces come from 512 TriviaQA sequences of 512 tokens, the last 20 held out. Targets use $D_\text{cold}{=}18$, $C_\text{cap}{=}36$, and a $D_\text{hot}$ under which the 98 farthest-reused experts of 128 receive a Hot target of 1. To test whether the scorer's OOD robustness depends on capacity, we sweep $h\in\{8,16,32,64,128\}$ and $d\in\{2,3,4\}$, each with 3 seeds, and measure hit rates on Qwen3-30B-A3B at $C{=}24$ under exact routing, averaged over 13 MoE layers (0--7, 15, 23, 31, 39, 47), on the 20 held-out TriviaQA sequences (ID) and 10 prompts of 512 tokens each from KoBEST/KLUE, MBPP, and GSM8K (OOD). Because trace, layers, and prompts differ from Table~\ref{tab:cache-policy}, Table~\ref{tab:scorer-sweep} reports hit rates relative to a retrained $h{=}32$, $d{=}3$ scorer.

\begin{table}[ht]
\centering
\small
\caption{Hit rate relative to the retrained $h{=}32$, $d{=}3$ scorer.}
\label{tab:scorer-sweep}
\resizebox{\linewidth}{!}{
\begin{tabular}{l|c|c|ccc}
\toprule
Policy & \# Params & TriviaQA (ID) & Korean & MBPP & GSM8K \\
\midrule
Retrained ($h{=}32$, $d{=}3$), 3 seeds     & 2{,}370        & 1.000 & 1.000 & 1.000 & 1.000 \\
Deployed scorer ($h{=}32$, $d{=}3$)        & 2{,}370        & 0.995 & 0.995 & 0.994 & 0.996 \\
Sweep range (15 configs $\times$ 3 seeds)  & 66 -- 67{,}074 & 0.998 -- 1.004 & 0.999 -- 1.001 & 0.999 -- 1.003 & 0.999 -- 1.002 \\
Belady (upper bound)                       & --             & 1.135 & 1.074 & 1.202 & 1.186 \\
\bottomrule
\end{tabular}
}
\end{table}

\section{Limitations}
\label{app:limitations}
CECAR is evaluated on three representative MoE LLMs and four standard benchmarks under a controlled on-device inference setup. While the results show consistent improvements across these settings, performance may vary under substantially different model architectures, workloads, or hardware platforms. Our testbed is a memory-constrained desktop GPU with SSD offloading and host DRAM pinned to $\sim$1~GB rather than a phone- or tablet-class SoC; we chose it because even the smallest evaluated model ($\sim$14~GB in BF16) exceeds phone-class memory, and because a mobile runtime would introduce confounds from the storage stack and power/thermal limits. Since CECAR's benefit derives from the latency gap between cached and non-resident experts, we expect it to be maintained or amplified on lower-bandwidth mobile storage such as UFS, but this is not verified here. Accuracy and throughput are measured on two different paths (Appendix~\ref{appendix:measurement}); accuracy is reported as mean $\pm$ SD over three decoding seeds and throughput as mean $\pm$ SD over repeated runs, whereas multi-seed variance for the hit rates in Table~\ref{tab:benchmark}, sensitivity to $D_\text{hot}$, $D_\text{cold}$ and $C_\text{cap}$, and a training-data-scale ablation remain to be reported. The architecture sweep in Appendix~\ref{appendix:scorer-sweep} shows that the scorer's out-of-distribution hit rate is unchanged across a $1000\times$ range of parameter counts, so its robustness is not an artifact of limited capacity; whether it is entirely due to the expert-independent input design has not been tested by ablating that design in isolation. Expert activation patterns are input-dependent, and we have not evaluated workloads with abrupt semantic shifts within a single request. Extending CECAR to batched serving, long-context generation, and quantized expert storage, and composing it with exact-routing prefetchers, remains future work.

\end{document}